\documentclass{egpubl}
\usepackage{pg2026}

\usepackage{comment}
\usepackage{amsmath}
\usepackage{amssymb}
\usepackage{multirow}
\usepackage[table,xcdraw]{xcolor}
\newcommand{\best}{red!40}
\newcommand{\second}{orange!40}
\newcommand{\third}{yellow!40}
\usepackage{booktabs}

\usepackage{subcaption}
\usepackage{bm}

\SpecialIssuePaper         % uncomment for final version of Computer Graphics Forum, special issue
\CGFStandardLicense
\usepackage[T1]{fontenc}
\usepackage{dfadobe}  

\usepackage{cite}  % comment out for biblatex with backend=biber
\BibtexOrBiblatex
\electronicVersion
\PrintedOrElectronic
\ifpdf \usepackage[pdftex]{graphicx} \pdfcompresslevel=9
\else \usepackage[dvips]{graphicx} \fi

\usepackage{url}
\usepackage{egweblnk}

\usepackage{multicol}
\usepackage{color}
\definecolor{myred}{RGB}{190,55,55}
\definecolor{myblue}{RGB}{45,95,170}
\definecolor{mygreen}{RGB}{45,170,15}
\definecolor{myorange}{RGB}{205,115,35}
\definecolor{mypurple}{RGB}{125,75,160}
\definecolor{myteal}{RGB}{35,135,145}
\definecolor{mybrown}{RGB}{145,90,55}

\title[EvoGS: Modeling Deformation Evolution for Dynamic Gaussian Splatting]
      {EvoGS: Modeling Deformation Evolution for Dynamic Gaussian Splatting}

\author[W. Dong et al.]
{\parbox{\textwidth}{\centering  
        Wei Dong\orcid{0000-0001-6109-5099},
        Shahram Shirani\orcid{0000-0002-7217-3467},
        Jun Chen\orcid{0000-0002-8084-9332},
        Han Zhou$^\dag$\orcid{0000-0001-7650-0755}
        }
        \\
{\parbox{\textwidth}{\centering The Department of Electrical \& Computer Engineering, McMaster University
       }
}
}
\usepackage{comment}
\usepackage{layout}
\usepackage{caption}
\begin{document}

% uncomment for using teaser
% \teaser{
%  \includegraphics[width=0.9\linewidth]{eg_new}
%  \centering
%   \caption{New EG Logo}
% \label{fig:teaser}
%}
{
\twocolumn[{
\renewcommand\twocolumn[1][]{#1}
\maketitle

\begin{center}
\setlength{\abovecaptionskip}{0.7mm}
    \includegraphics[width = 0.96\textwidth]{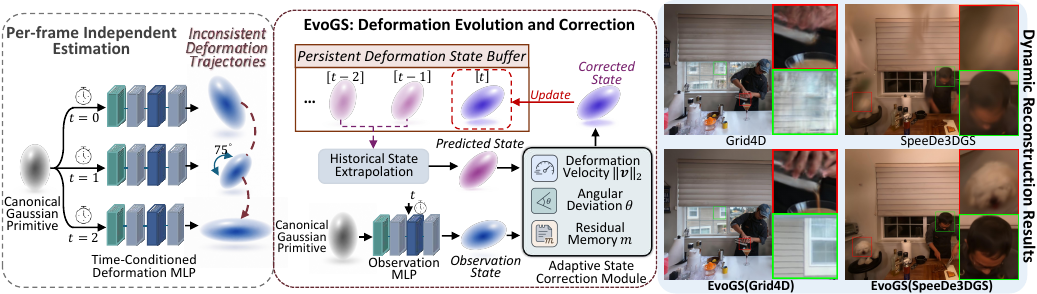}
    \captionof{figure}{Existing dynamic 3DGS methods often estimate deformations separately at each timestamp, which can lead to temporal artifacts and unstable reconstruction. EvoGS maintains persistent deformation states and performs evolution-based correction. The qualitative results show that integrating EvoGS improves different dynamic Gaussian backbones, producing sharper details and better-preserved geometry.}
    \vspace{7mm}
    \label{fig1}
\end{center}}]
}

%\maketitle
%-------------------------------------------------------------------------

\begin{abstract}
Recent extensions of 3D Gaussian Splatting (3DGS) enable real-time novel view synthesis in dynamic scenes by learning time-conditioned Gaussian deformations. However, existing MLP-based methods typically estimate deformations independently at each timestamp, making them less robust to large or abrupt motions.
To address this issue, we propose \textbf{EvoGS}, a 3DGS-based dynamic reconstruction framework that models Gaussian deformation as a temporal evolution process. EvoGS maintains persistent deformation states for each Gaussian, extrapolates future states from historical deformation states, and corrects the predictions with MLP-derived observations. The correction is adaptively weighted using a temporal residual memory and evolution statistics such as deformation velocity and trajectory deviation.
To further improve reconstruction quality, EvoGS introduces deformation-aware densification. Clone and split operations are performed along corrected deformation directions, while an uncertainty-aware strategy suppresses densification for Gaussians with unstable deformation histories. Experiments show that EvoGS improves dynamic novel view synthesis quality and achieves competitive performance across benchmarks.

\begin{CCSXML}
<ccs2012>
<concept>
<concept_id>10010147.10010371.10010372</concept_id>
<concept_desc>Computing methodologies~Rendering</concept_desc>
<concept_significance>500</concept_significance>
</concept>
<concept>
<concept_id>10010147.10010371.10010352</concept_id>
<concept_desc>Computing methodologies~Computer graphics</concept_desc>
<concept_significance>500</concept_significance>
</concept>
<concept>
<concept_id>10010147.10010257.10010293.10010294</concept_id>
<concept_desc>Computing methodologies~Neural networks</concept_desc>
<concept_significance>300</concept_significance>
</concept>
</ccs2012>
\end{CCSXML}

\ccsdesc[500]{Computing methodologies~Rendering}
\ccsdesc[500]{Computing methodologies~Computer graphics}
\ccsdesc[500]{Computing methodologies~Neural networks}

\printccsdesc

\end{abstract}  

\def\thefootnote{\dag}\footnotetext{Corresponding author}

\section{Introduction}
As a fundamental task in computer vision, dynamic 3D scene reconstruction is to reconstruct and render scenes involving object motion with temporal consistency, geometric accuracy, and high visual fidelity, which has wide applications including augmented reality\/virtual reality (AR$\/$VR), autonomous navigation, and commercial advertising. Recent years have witnessed significant progress driven by Neural Radiance Fields (NeRF~\cite{nerf}) and its extensive variants~\cite{du2021neural,gao2021dynamic,park2021hypernerf,park2021nerfies,D-NeRF,fang2022fast}, which model scenes using time-conditioned MLPs to represent geometry and appearance over time. While powerful in capturing complex motion, these methods are notoriously slow due to the cost of volumetric ray marching and implicit representation, rendering them unsuitable for real-time applications.

\begin{figure*}[!h]
    \centering
    \setlength{\abovecaptionskip}{2mm}
    \includegraphics[width=0.85\textwidth]{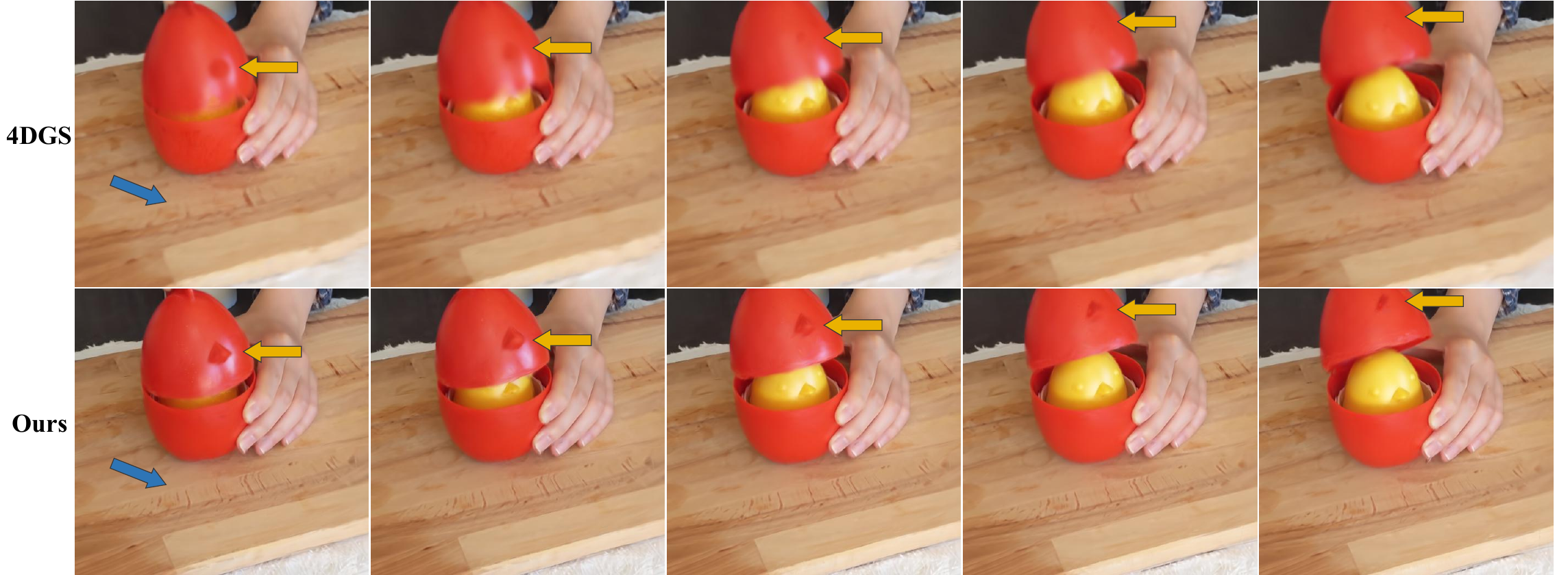}
    \caption{Under a fixed camera pose with varying timesteps ($\Delta t = 0.01$), the baseline 4DGS~\cite{4DGS} produces blurry and temporally inconsistent results, where \textit{the shell shape gradually fades across frames}. In the baseline result, the colored arrows highlight typical artifacts: yellow indicates geometric distortion, while blue marks blurred wood textures.In contrast, EvoGS preserves sharper details, more stable geometry, and temporally coherent appearances in dynamic regions through persistent temporal deformation evolution. This evolution process also indirectly reduces ambiguity in static regions by mitigating motion-induced inconsistencies, leading to better overall reconstruction.}
    \label{fig_consistency}
    \vspace{-3mm}
\end{figure*}

 To address the efficiency bottleneck, 3D Gaussian Splatting (3DGS~\cite{3Dgaussian}) has emerged as a new paradigm for real-time rendering with high-quality results. By representing the scene as a collection of textured 3D Gaussians projected onto the image plane, 3DGS allows for real-time view synthesis~\cite{3Dgaussian, sun20243dgstream, katsumata2024compact, han2025litags}. Recent extensions of 3DGS into dynamic settings attempt to model scene deformation over time by learning per-Gaussian offsets using MLPs. Notable works such as 4D-GS~\cite{4DGS}, SC-GS~\cite{SC-GS}, D-3DGS~\cite{deformablegs}, MotionGS~\cite{motiongs}, Grid4D~\cite{grid4d}, DASH~\cite{dash}, MoDec-GS~\cite{modecgs} adopt frame-wise regression networks to estimate deformation at each timestep (Fig.~\ref{fig1}). However, these methods typically treat each frame independently, without modeling temporal context in the input sequences. This unavoidably results in inconsistent deformation trajectories across frames and perceptual flickering in the rendered video sequences (see Fig.~\ref{fig_consistency}).

To address this issue, we propose \textbf{EvoGS}, a dynamic 3D \textbf{G}aussian \textbf{S}platting framework designed to model the temporal \textbf{EVO}lution of Gaussian deformations. Unlike existing MLP-based approaches that query the deformation of each timestamp independently, EvoGS treats deformation estimation as an \textit{evolving state modeling problem}. For each Gaussian primitive, we maintain a persistent \textbf{deformation state buffer} that records corrected deformation states across training timesteps. Given historical states, EvoGS extrapolates a predicted deformation state for the current timestep and further refines it with an observation derived from a deformation MLP. The correction is adaptively weighted according to a \textbf{temporal residual memory and motion evolution statistics}, such as deformation velocity and angular deviation, allowing the model to balance historical prediction and current observation. Furthermore, EvoGS leverages the corrected deformation states for deformation-aware densification, where clone and split operations are performed along corrected deformation directions instead of relying only on local perturbations around the parent Gaussian. To avoid expanding unreliable primitives, we also introduce an uncertainty-aware densification criterion that suppresses densification for Gaussians exhibiting unstable deformation evolution.

Our contributions are summarized as follows:

$\diamond$ We propose \textbf{EvoGS}, a 3DGS-based framework that models Gaussian deformation as a temporal evolution process for dynamic scene reconstruction.

$\diamond$ We introduce persistent deformation state modeling, where historical corrected states are stored, extrapolated, and corrected using MLP-derived observations for more reliable estimation.

$\diamond$ We develop an adaptive correction strategy driven by temporal residual memory and deformation evolution statistics, including deformation velocity and angular deviation.

$\diamond$ We propose deformation-aware densification, which guides clone/split operations using corrected deformations and suppresses densification for Gaussians with unstable deformation histories.

$\diamond$ Extensive experiments demonstrate that EvoGS improves dynamic view synthesis quality across multiple benchmarks.
\section{Related Works}
\label{sec:rela}

\textbf{Dynamic NeRF.}
Dynamic novel view synthesis remains challenging due to temporal variation in input images. The success of NeRF~\cite{nerf} in representing static scenes with implicit neural fields has inspired its adaptation to dynamic settings. Some methods incorporate temporal information implicitly through time-conditioned inputs or latent embeddings~\cite{du2021neural,gao2021dynamic}, while others~\cite{park2021hypernerf,park2021nerfies,D-NeRF,fang2022fast, nerf1} learn explicit deformation fields that transform timestamped 3D points to a canonical space. Additional approaches separate scenes into static and dynamic parts~\cite{song2023nerfplayer}, apply 4D grid-based structures~\cite{fridovich2023k,li2022streaming,wang2023mixed,cao2023hexplane} or use keyframes for redundancy reduction~\cite{li2022neural,attal2023hyperreel}. Despite progress, spatial-temporal coupling and the computational overhead of temporal modeling remain fundamental challenges.
\begin{figure}[t]
\setlength{\abovecaptionskip}{1mm}
\setlength{\parskip}{0mm} % Ensures no extra paragraph spacing is added
\setlength{\baselineskip}{0mm} % Reduces line spacing for finer control
\centering
    \includegraphics[width=0.9\linewidth]{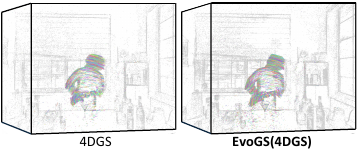}
\caption{Qualitative comparison of Gaussian deformation trajectories. The baseline 4DGS exhibits scattered motion traces with local jitter, crossings, and outlier trajectories around the moving subject.In contrast, EvoGS(4DGS) yields more compact and structure-aligned trajectories over time.The reduced trajectory dispersion suggests more stable deformation evolution. Notably, only Gaussians satisfying the visibility, opacity, motion-magnitude, and trackability criteria are visualized.  } %

\label{fig_trajectory}
\vspace{-2mm}
\end{figure}

\begin{figure*}[t]
    \centering
    \includegraphics[width=0.9\textwidth]{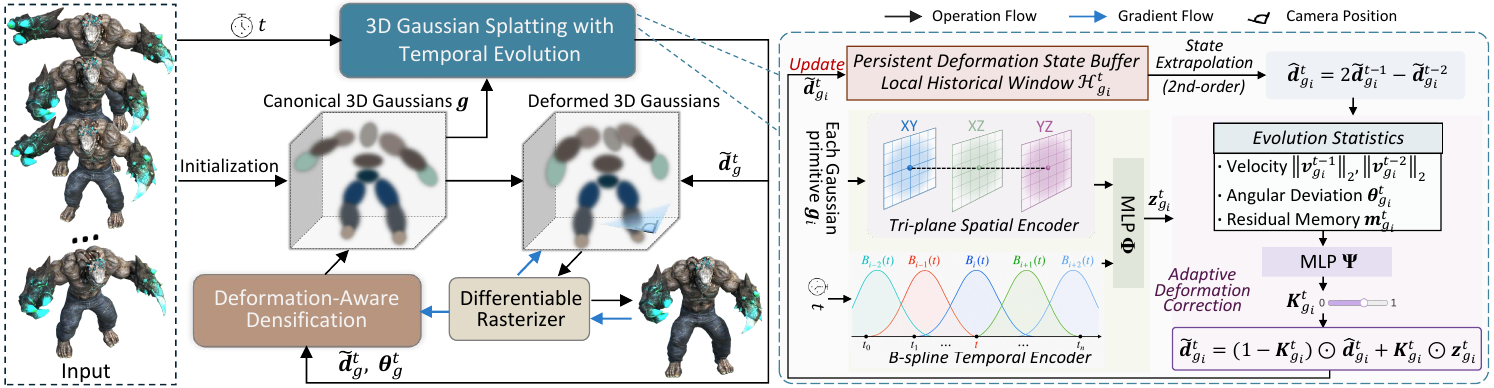}
    \caption{The overall architecture of \textbf{EvoGS}. Instead of the standard modeling for Gaussian deformation using observation MLP, we propose the temporal deformation evolution mechanism to temporally learn Gaussian deformation field across time. Furthermore, we develop the deformation-aware densification strategy, which initializes newly created Gaussians based on the updated deformation and prevents densification in regions with high deformation uncertainty.}
    \label{fig_overall_framework}
    \vspace{-4mm}
\end{figure*}

\noindent  \textbf{Dynamic Gaussian Splatting.} 
Gaussian Splatting (GS)~\cite{3Dgaussian} excels in rendering efficiency and fidelity for complex scenes. Recent advances have extended 3DGS for dynamic scenes~\cite{gs-1, gs-2, gs-3, gs-4, gs-5, deformablegs, 4DGS, hexplane, SC-GS, gs1, gs2, gs3, gs4, gs5, gs6, gs7}. Yang \textit{et al.}~\cite{deformablegs} employ an MLP to generate a time-dependent deformation field over the canonical Gaussian space. Wu \textit{et al.}~\cite{4DGS} propose a spatial-temporal encoder with multi-resolution HexPlanes~\cite{hexplane} and a compact MLP to learn Gaussian deformation. SC-GS~\cite{SC-GS} predict time-varying transformations of sparse control points via an MLP, whose interpolated effects drive Gaussian motion. E-D3DGS~\cite{Per-Gaussian} employs per-Gaussian and temporal embeddings for fine-grained motion modeling. While these methods can model frame-wise motion, they typically rely on temporally conditioned networks without modeling motion dynamics across frames, which may lead to temporal inconsistencies in complex scenes. MotionGS~\cite{motiongs} addresses this via a motion-guided Gaussian deformation framework supervised by decoupled optical flow. However, its reliance on externally estimated optical flow limits its ability to robustly capture consistent motion under uncertain or ambiguous dynamics.

\noindent \textbf{Temporal State-Space Modeling.}
State-space models and Kalman-inspired neural networks provide a principled perspective for modeling temporal evolution through recursive prediction and correction. Early deep Kalman-style models combine latent state-space formulations with deep generative or discriminative networks~\cite{krishnan2015deep, haarnoja2016backprop, fraccaro2017disentangled}, enabling end-to-end learning of temporal dynamics. KalmanNet~\cite{becker2021kalmannet} preserves the recursive structure of classical Kalman filtering while learning adaptive gains from data, and Horvath et al.~\cite{horvath2023dkfproof} further analyze the approximation ability of deep Kalman filters in non-Markovian settings. Given that directly employing image restoration methods \cite{codeformer, glare, gppllie, ecmamba, varlide} to video contents introduces inconsistency across frames, KEEP~\cite{keep} introduces a Kalman-inspired method for stable video face restoration. More recently, KFD-NeRF~\cite{kf-nerf} introduces a Kalman-style update into dynamic NeRF by using locally linear motion prediction to guide deformation learning. Although these methods demonstrate the benefit of recursive state modeling, their formulations are not directly designed for Gaussian primitives in dynamic 3DGS, where each Gaussian undergoes time-varying deformation and densification. In contrast, EvoGS models Gaussian deformation as a persistent temporal evolution process. Instead of independently regressing deformations at each timestamp, EvoGS maintains per-Gaussian deformation states, extrapolates future states from historical trajectories, and adaptively corrects them using MLP-derived observations and evolution statistics.
\section{Preliminary}
\subsection{Radiance Modeling via 3D Gaussian Splatting}
\label{sec_prelimninary_3DGS}

3D Gaussian Splatting (3DGS)~\cite{3Dgaussian} represents a scene using a set of spatially continuous, anisotropic volumetric primitives $\mathcal{G} = \{\bm{g}_i | i=1,...,N \}$, where $N$ represents the number of Gaussian primitives. Each point $\bm{g}_j$ in the scene is modeled as a Gaussian blob parameterized by: $\bm{g}_i = \left\{ \bm{\mu}_i \in \mathbb{R}^3, \ \bm{s}_j \in \mathbb{R}^{3}, \ \bm{q}_i \in \mathbb{R}^{4}, \ \sigma_i \in (0,1), \ \bm{c}_j \in \mathbb{R}^3 \right\}$,
where $\bm{\mu}_i$ is the 3D center, $\bm{s}_i$ is the scale vector encoded as a diagonal covariance, $\bm{q}_i$ is the quaternion representing rotation, $\sigma_i$ is the opacity, and $\bm{c}_j$ denotes its spherical harmonics-encoded color. Notably, the scale and rotation jointly determine the 3D covariance matrix of each Gaussian
%\begin{equation}
   $ \bm \Sigma_i = \bm{q}_i\bm{s}_i\bm{s}_i^{T}\bm{q}_i^{T}$.
%\end{equation}

For rendering, each $\bm{g}_i$ is projected into the 2D space using the following 2D covariance matrix $\bm \Sigma^{\prime}$:
%\begin{equation}
  $   \bm \Sigma_i^{\prime} = \bm{JV\Sigma}_i \bm{V}^T\bm{J}^T,$ 
%\end{equation}
where $J$ is the Jacobian of the affine approximation of the projective transformation, $V$ symbolizes the view matrix, transitioning from world to camera coordinates. Then, the resulting 2D ellipsoids are rasterized and composited via alpha blending based on depth ordering:
\vspace{-2mm}
\begin{equation}
\begin{split}
    \bm{C}(\bm{p}) &= \sum_{i=1}^{N} T_i \alpha_i \bm{c}_i,  \quad T_i = \Pi_{j=1}^{i - 1}(1 - \alpha_{j}),\\
    &\alpha_i = \sigma_i e^{-\frac{1}{2} (\bm{p} - \mu_i^{\prime})^T \sum^{\prime} (\bm{p}  - \mu_i^{\prime}) },
\end{split}
\end{equation}

\noindent where $\bm{C}(\bm{p})$ denotes the rendered color of pixel $\bm{p}$ on the image plane, and $\mu_i^{\prime}$ represents the coordinates of 3D Gaussian center projected onto the 2D image plane. This pipeline circumvents volumetric ray integration and enables real-time differentiable rendering.

\subsection{Temporal State Prediction and Correction}

Temporal state-space models provide a natural framework for estimating a latent state from historical predictions and current observations. A classical example is the Kalman filter, which considers a linear state-space model:
\begin{equation}
\bm{x}_{t} = \bm{F}\bm{x}_{t-1} + \bm{w}_{t}, \quad
\bm{z}_{t} = \bm{H}\bm{x}_{t} + \bm{m}_{t},
\label{eq:kalman_linear_state}
\end{equation}
where $\bm{x}_{t}$ is the latent state, $\bm{z}_{t}$ the observation, $\bm{F}$ the state-transition matrix, $\bm{H}$ the observation matrix, and $\bm{w}_{t}\!\sim\!\mathcal{N}(\bm{0},\bm{Q})$, $\bm{m}_{t}\!\sim\!\mathcal{N}(\bm{0},\bm{R})$ denote process and measurement noise with covariances $\bm{Q}$ and $\bm{R}$, respectively. The latent state is recursively estimated through a prediction--correction procedure:
\begin{equation}
\hat{\bm{x}}_{t}^{-} = \bm{F}\hat{\bm{x}}_{t-1}, \qquad
\hat{\bm{x}}_{t} = \hat{\bm{x}}_{t}^{-} + \bm{K}_{t}\big(\bm{z}_{t} - \bm{H}\hat{\bm{x}}_{t}^{-}\big),
\label{eq:kalman_update}
\end{equation}
with $\bm{H}=\bm{I}$, the update becomes 
$\hat{\bm{x}}_{t} = (\bm{I}-\bm{K}_{t})\hat{\bm{x}}_{t}^{-} + \bm{K}_{t}\bm{z}_{t}$, 
which shows that the correction step adaptively interpolates between the historical prediction and the current observation.

In this work, we use this recursive prediction-correction principle as a conceptual basis for modeling Gaussian deformation evolution instead of implementing a classical Kalman filter or propagating covariance matrices. This is because explicitly maintaining per-Gaussian covariance matrices would introduce substantial memory and computational overhead for the large and dynamically changing Gaussian set, while reliable process and observation covariances are also difficult to define for nonlinear, non-rigid deformations.

\section{Method}
\label{sec:method}

The overall framework of EvoGS is illustrated in Fig.~\ref{fig_overall_framework}. Specifically, EvoGS models dynamic Gaussian deformation as a temporal evolution process rather than independent per-timestamp regression, commonly used in existing approaches. For each Gaussian, EvoGS maintains persistent deformation states, extrapolates the current deformation from historical corrected states, and adaptively refines it using an MLP-derived observation. Sec.~\ref{sec:deform_evolution} introduces the persistent deformation buffer and the historical state extrapolation used to obtain the evolution prior. Sec.~\ref{sec:deform_correction} describes the observation network with tri-plane spatial and B\mbox{-}spline temporal encoders, as well as the adaptive correction module based on innovation memory and motion evolution statistics. Furthermore, we develop the uncertainty-aware Gaussians control strategy to adjust the density of Gaussians for areas that exhibits with high motion uncertainty in Sec.~\ref{sec_densification}. Moreover, different from the original Gaussians clone and split mechanism proposed in 3DGS~\cite{3Dgaussian}, we propose the deformation-aware clone and split process for these new instantiated Gaussians. The optimization details of our method are discussed in Sec.~\ref{sec_optimization}.
\subsection{Deformation Evolution with Persistent States}
\label{sec:deform_evolution}

Given a canonical 3D Gaussian primitive $\bm{g}_i$, dynamic 3DGS methods usually predict its time-dependent deformation with a time-conditioned network. Instead of treating the deformation at each timestamp as an independent network output, EvoGS models Gaussian deformation as a persistent temporal state. We denote the deformation state of $\bm{g}_i$ at timestamp $t$ as $\bm{d}_i^t \in \mathbb{R}^{D}$, which can represent position, scale, rotation, or other deformation offsets depending on the adopted dynamic 3DGS backbone.

\noindent \textbf{Persistent Deformation States.}
For each Gaussian primitive $\bm{g}_i$, EvoGS maintains a persistent deformation buffer
\begin{equation}
\mathcal{B}_{g_i} = \{\tilde{\bm{d}}_{g_i}^1, \tilde{\bm{d}}_{g_i}^2, ..., \tilde{\bm{d}}_{g_i}^T\},
\label{eq_buffer_deformation}
\end{equation}
where $\tilde{\bm{d}}_{g_i}^t$ denotes the corrected deformation state of $\bm{g}_i$ at timestamp $t$. Given a target timestamp $t$, EvoGS queries a local historical window from this buffer:
\begin{equation}
\mathcal{H}_{g_i}^t =
\{
\tilde{\bm{d}}_{g_i}^{t-1},
\tilde{\bm{d}}_{g_i}^{t-2}
\}
\subset \mathcal{B}_{g_i},
\label{eq_buffer_deformation_window}
\end{equation}
which provides a persistent motion history for each Gaussian and serves as the basis for predicting its current deformation. After the correction step in Sec.~\ref{sec:deform_correction}, the updated state $\tilde{\bm{d}}_{g_i}^{t}$ is written back to the buffer and reused for subsequent timestamps.

\noindent\textbf{Historical State Evolution.}
Given the local historical window $\mathcal{H}_i^t$, EvoGS predicts an evolution prior for the current deformation state from the corrected historical states. We adopt a second-order temporal extrapolation:
\begin{equation}
\hat{\bm{d}}_{g_i}^{t}
=
2\tilde{\bm{d}}_{g_i}^{t-1}
-
\tilde{\bm{d}}_{g_i}^{t-2},
\label{eq:state_evolution}
\end{equation}
where $\hat{\bm{d}}_i^{t}$ denotes the predicted deformation state before incorporating the current observation. This design explicitly propagates the historical motion trend stored in the persistent deformation buffer. Since complex non-rigid motion may deviate from this extrapolated prior, EvoGS further corrects it using the current MLP-derived observation through the adaptive correction module in Sec.~\ref{sec:deform_correction}.

\begin{figure*}[!t]
    \centering
    \setlength{\abovecaptionskip}{2mm}
    \includegraphics[width=1\textwidth]{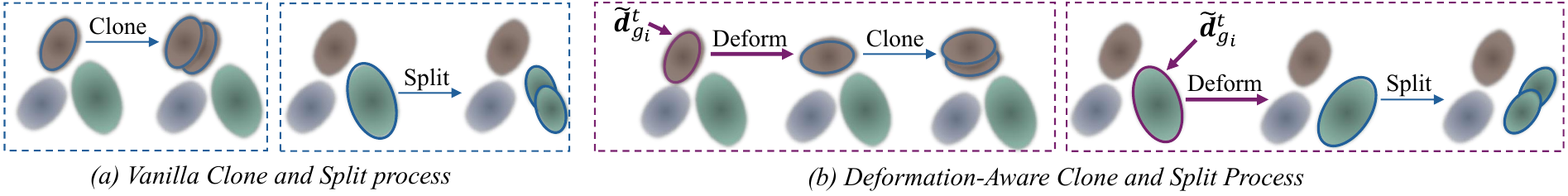}
    \caption{Illustration of the proposed deformation-aware clone-and-split process. Compared with the vanilla strategy, our method integrates deformation cues to initialize new Gaussians with motion-adaptive geometry, leading to more consistent and stable updates across frames.}
    \label{fig_densify}
    \vspace{-2mm}
\end{figure*}

\textbf{Discussion.} The second-order extrapolation is used only as a lightweight short-horizon prior, rather than as the final deformation estimate or a strict constant-velocity motion assumption. When the motion deviates from this prior, especially under non-rigid or sudden changes, the current MLP-derived observation is used to correct the prediction through the adaptive correction module. Therefore, the extrapolation provides temporal continuity, while the observation branch compensates for deviations from the local motion trend. The current formulation assumes approximately uniform temporal spacing, as used by the evaluated benchmarks.

\subsection{Adaptive Deformation Correction}
\label{sec:deform_correction}

The historical state evolution in Sec.~\ref{sec:deform_evolution} provides an evolution prior $\hat{\bm{d}}_i^t$ for each Gaussian primitive. However, this extrapolated prior may become inaccurate under abrupt or non-rigid motion. To account for such deviations, EvoGS further corrects the predicted state using a deformation observation estimated from the current timestamp.

\noindent\textbf{Observation Network with Tri-plane Spatial and B\mbox{-}spline Temporal Encoders.}
We use an observation network $\boldsymbol{\Phi}$ to estimate the current deformation observation:
\begin{equation}
\bm{z}_{g_i}^t = \boldsymbol{\Phi}(\bm{g}_i,t),
\label{eq}
\end{equation}
where $\bm{z}_{g_i}^t$ denotes the MLP-derived deformation observation for Gaussian $\bm{g}_i$ at timestamp $t$. Following the dynamic 3DGS backbone, $\boldsymbol{\Phi}$ predicts the same type of deformations. To improve efficiency and temporal smoothness, we parameterize $\boldsymbol{\Phi}$ with tri-plane spatial features and a B\mbox{-}spline temporal encoder. Specifically, spatial features are sampled from the XY, XZ, and YZ feature planes according to the canonical Gaussian position, while the timestamp is encoded by a uniform cubic B\mbox{-}spline basis. The sampled spatial features and temporal embedding are then decoded by a lightweight MLP to produce $\bm{z}_i^t$. Compared with directly using sinusoidal positional encodings and a large MLP, this design preserves local spatial structure and provides smooth temporal interpolation.

\noindent\textbf{Evolution-aware Correction Weight.}
The reliability of the extrapolated prior and the current observation varies across Gaussians and timestamps. EvoGS therefore predicts an adaptive correction weight from temporal evolution statistics.

\noindent \textit{\textbf{(1) Temporal Residual Memory.}} We first compute the prediction-observation residual: $\bm{r}_{g_i}^t = \bm{z}_{g_i}^t - \hat{\bm{d}}_{g_i}^t.$ To make the correction stable over time, we maintain a temporal residual memory:
\begin{equation}
\bm{m}_{g_i}^t =
\beta \bm{m}_{g_i}^{t-1} + (1-\beta)\bm{r}_i^t,
\label{eq}
\end{equation}
where $\beta$ is a momentum coefficient. This memory records recent prediction uncertainty and prevents the correction from being dominated by a single noisy observation.

\noindent \textit{\textbf{(2) Observation Instability.}} Based on the positional component in the persistent deformation buffer $\tilde{\bm{d}}_{g_{i,x}}^{t-3}, \tilde{\bm{d}}_{g_{i,x}}^{t-2}, \tilde{\bm{d}}_{g_{i,x}}^{t-1}$, we compute the magnitudes of deformation velocities across successive frames, ($\|\bm{v}_{g_i}^{t-1}\|_2$ and $\|\bm{v}_{g_i}^{t-2}\|_2$), as defined in Eq.~\ref{eq:proxy_vel}, where large temporal displacements suggest abrupt motion and reduced predictive reliability. 

\noindent \textit{\textbf{(3) Angular Deviation.}} We further calculate the angle between adjacent motion vectors, given in Eq.~\ref{eq_proxy_angle}. Significant directional changes imply motion irregularity and are treated as proxies for elevated uncertainty. 
\begin{align}
    \|\bm{v}_{g_i}^{\tau}\|_2 &= \|\,\tilde{\bm{d}}_{g_{i,x}}^{\tau}-\tilde{\bm{d}}_{g_{i,x}}^{\tau-1}\,\|_2, \qquad \tau \in \{t{-}1,\, t{-}2\}, 
    \label{eq:proxy_vel} \\
     &\theta_{g_i}^t = \arccos \left( \frac{\langle \bm{v}_{g_i}^{t-2},\, \bm{v}_{g_i}^{t-1} \rangle}{\|\bm{v}_{g_i}^{t-2}\|\cdot \|\bm{v}_{g_i}^{t-1}\| + \epsilon} \right),\
    \label{eq_proxy_angle}
\end{align}
where $\epsilon$ is set to $10^{-6}$ for numerical stability. These uncertainty-related representations are fed to a lightweight MLP $\bm{\Psi}$ to estimate the gain: $\bm{K}_{g_i}^{t} = \bm \Psi(\bm{m}_{g_i}^t, \|\bm{v}_{g_i}^{t-2}\|_2, \|\bm{v}_{g_i}^{t-1}\|_2,  \theta_{g_i}^t)$. 
$\bm{\Psi}$ adopts a three-layer MLPs with hidden size 32 and ReLU activations. A final sigmoid activation in $\bm{\Psi}$ ensures the predicted gain lies within $[0, 1]$.
The final corrected deformation state is obtained by interpolating between the extrapolated prior and the current observation:
\begin{equation}
\tilde{\bm{d}}_{g_i}^t = (1-\bm{K}_{g_i}^t)\odot \hat{\bm{d}}_{g_i}^t + \bm{K}_{g_i}^t \odot \bm{z}_{g_i}^t ,
\label{eq:adaptive_correction}
\end{equation}
where $\odot$ denotes element-wise multiplication. The corrected state $\tilde{\bm{d}}_{g_i}^t$ is used for rendering and written back to the deformation buffer for subsequent state evolution. For the first three timesteps, we skip Eq.~\ref{eq:adaptive_correction} and use the extrapolated prior directly.

\subsection{Deformation-adaptive Control of Gaussians}
\label{sec_densification}

To maintain representation fidelity and stabilize optimization in dynamic scenes, we introduce a deformation-adaptive densification module with two components: (i) an uncertainty-aware trigger that avoids densifying unstable Gaussians; (ii) a deformation-aware clone/split strategy that aligns newly created Gaussians with the underlying motion.

\begin{figure*}[h]
  \centering

  \begin{subfigure}{0.192\linewidth}%
    \centering
    \includegraphics[width=\linewidth]{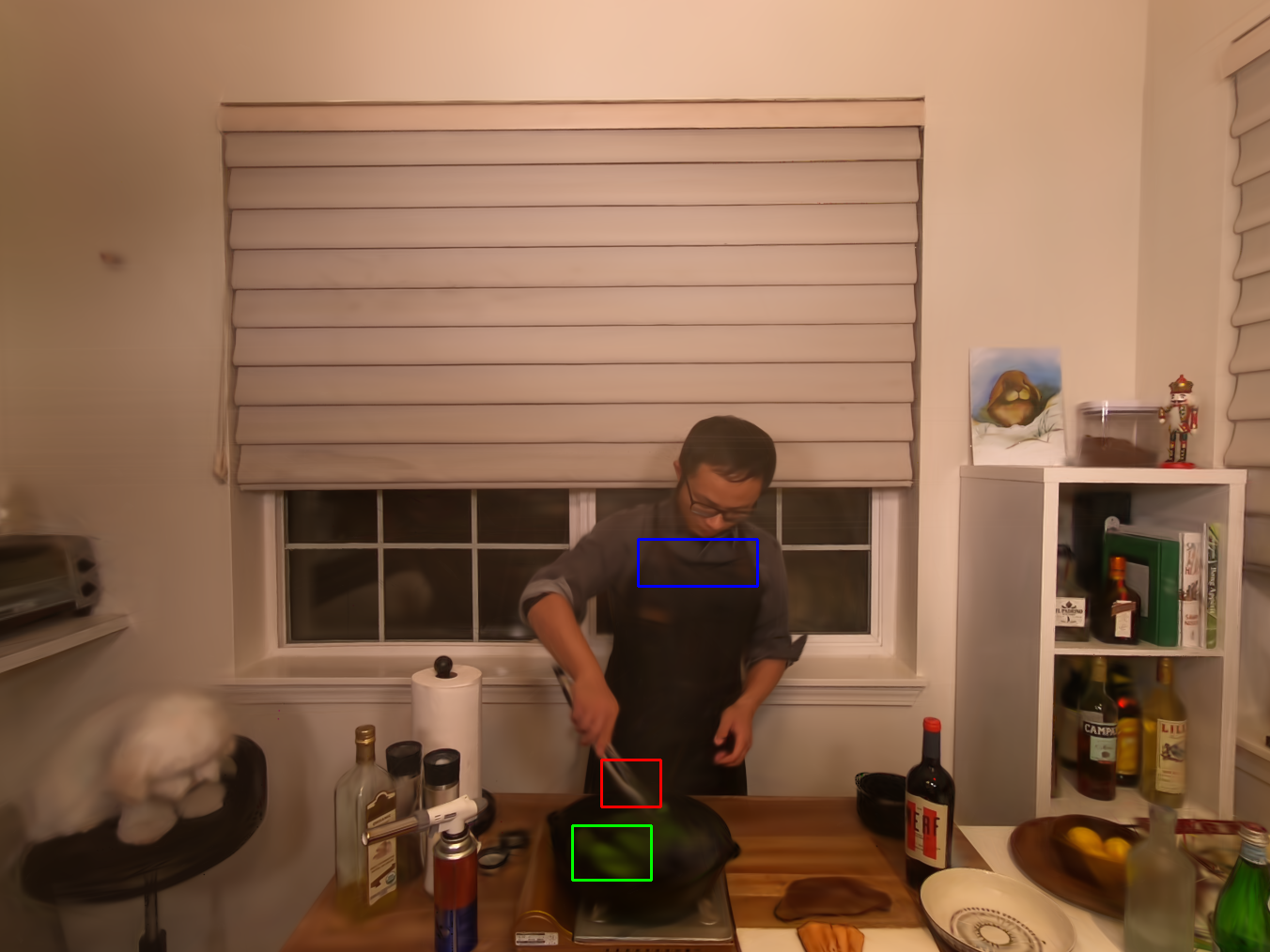}\\[-1pt]
    
    \begin{minipage}[b]{0.47\linewidth}
      \includegraphics[width=\linewidth]{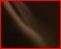}
    \end{minipage}%
    \begin{minipage}[b]{0.538\linewidth}
      \includegraphics[width=\linewidth]{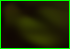}
    \end{minipage}\\[-1pt]
    
    \includegraphics[width=\linewidth]{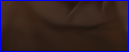}
    \caption*{4DGS}
  \end{subfigure}
  \hfill
  \begin{subfigure}{0.192\linewidth}%
    \centering
    \includegraphics[width=\linewidth]{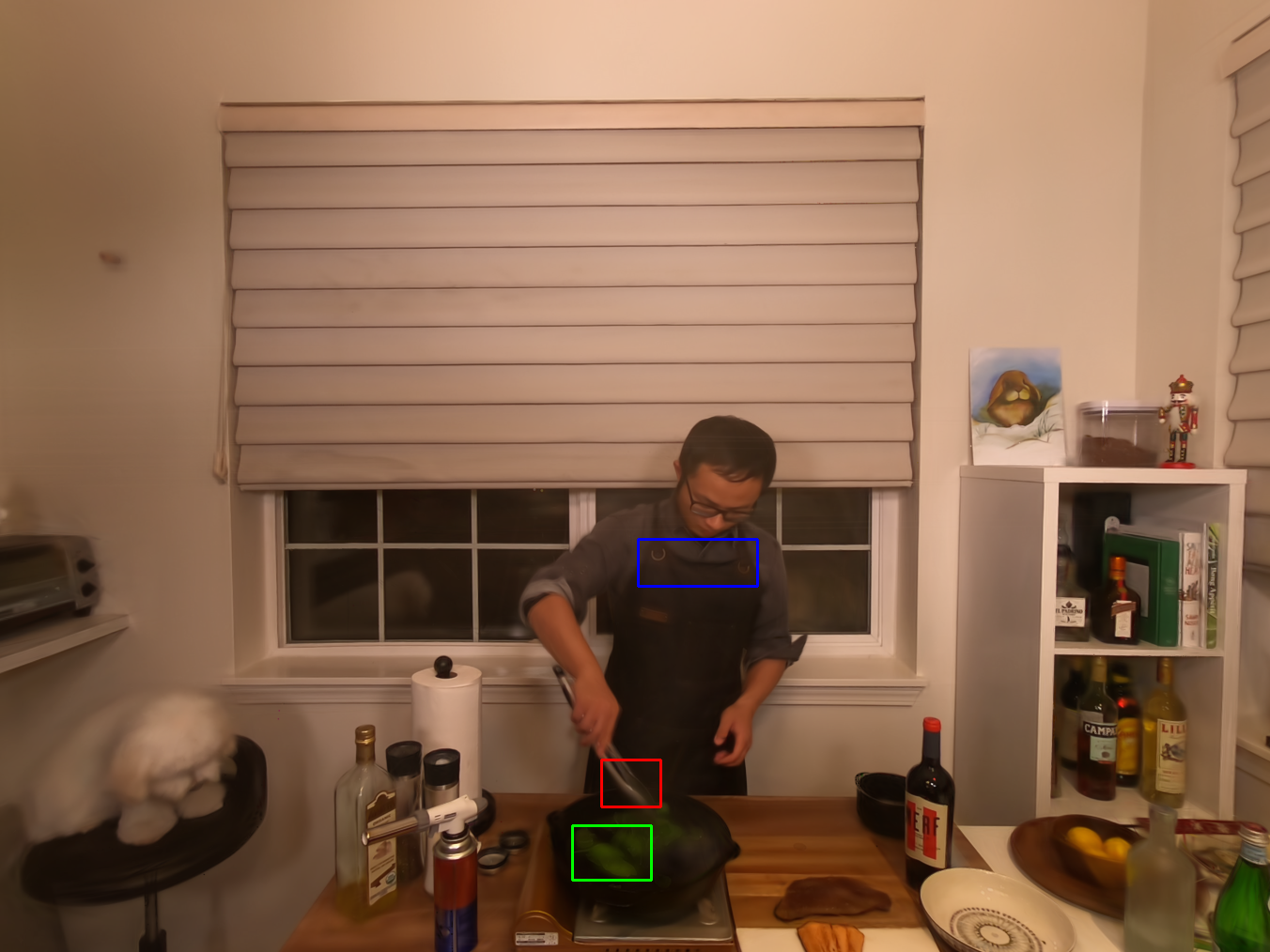}\\[-1pt]
    
    \begin{minipage}[b]{0.47\linewidth}
      \includegraphics[width=\linewidth]{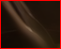}
    \end{minipage}%
    \begin{minipage}[b]{0.538\linewidth}
      \includegraphics[width=\linewidth]{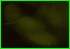}
    \end{minipage}\\[-1pt]
    
    \includegraphics[width=\linewidth]{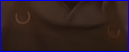}
    \caption*{\textbf{EvoGS(4DGS)}}
  \end{subfigure}
  \hfill
  \begin{subfigure}{0.192\linewidth}%
    \centering
    \includegraphics[width=\linewidth]{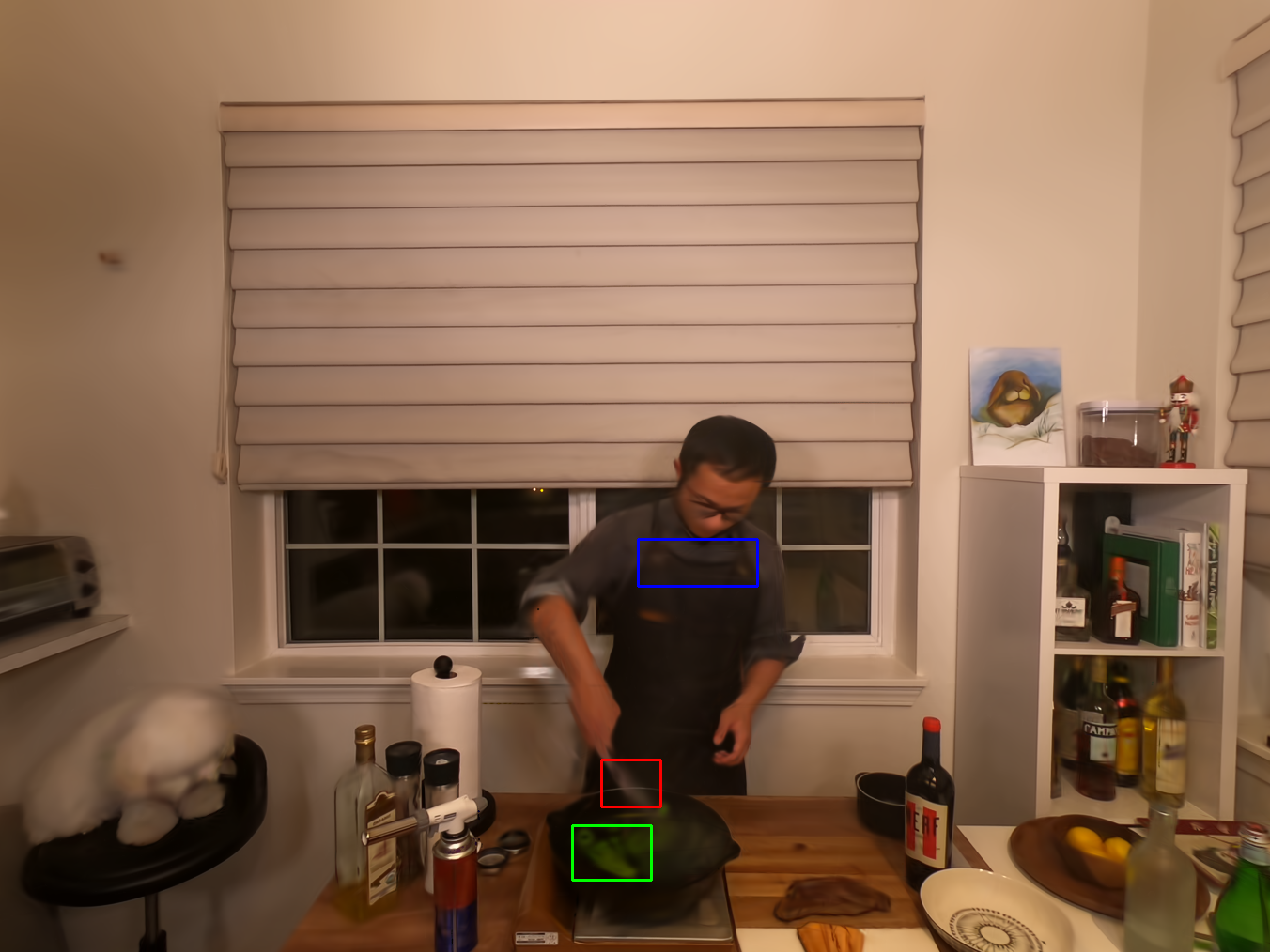}\\[-1pt]
    
    \begin{minipage}[b]{0.47\linewidth}
      \includegraphics[width=\linewidth]{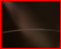}
    \end{minipage}%
    \begin{minipage}[b]{0.538\linewidth}
      \includegraphics[width=\linewidth]{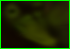}
    \end{minipage}\\[-1pt]
    
    \includegraphics[width=\linewidth]{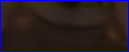}
    \caption*{DASH}
  \end{subfigure}
  \hfill
  \begin{subfigure}{0.192\linewidth}%
    \centering
    \includegraphics[width=\linewidth]{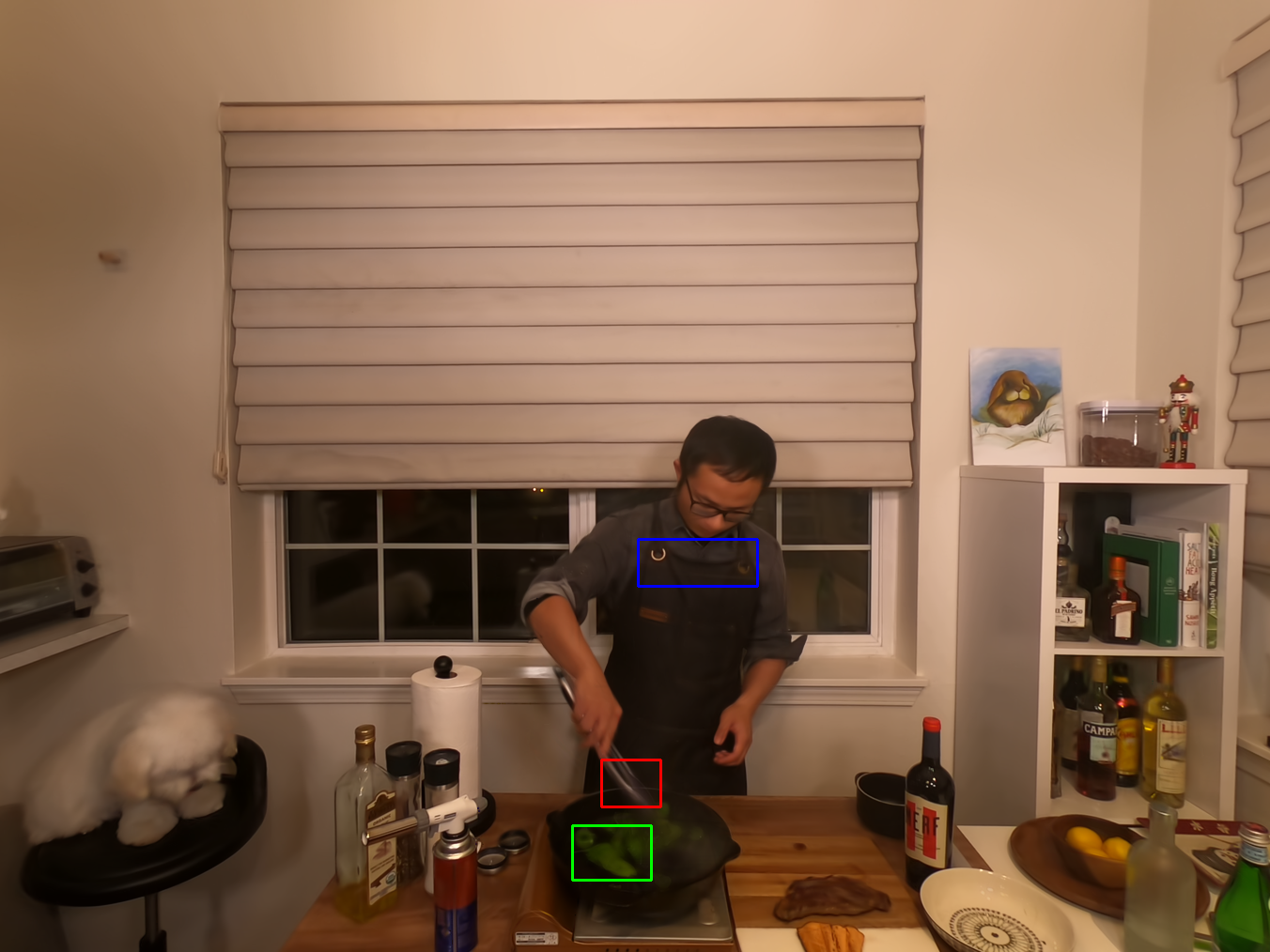}\\[-1pt]
    
    \begin{minipage}[b]{0.47\linewidth}
      \includegraphics[width=\linewidth]{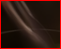}
    \end{minipage}%
    \begin{minipage}[b]{0.538\linewidth}
      \includegraphics[width=\linewidth]{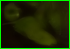}
    \end{minipage}\\[-1pt]
    
    \includegraphics[width=\linewidth]{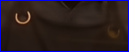}
    \caption*{\textbf{EvoGS(DASH)}}
  \end{subfigure}
  \hfill
  \begin{subfigure}{0.192\linewidth}%
    \centering
    \includegraphics[width=\linewidth]{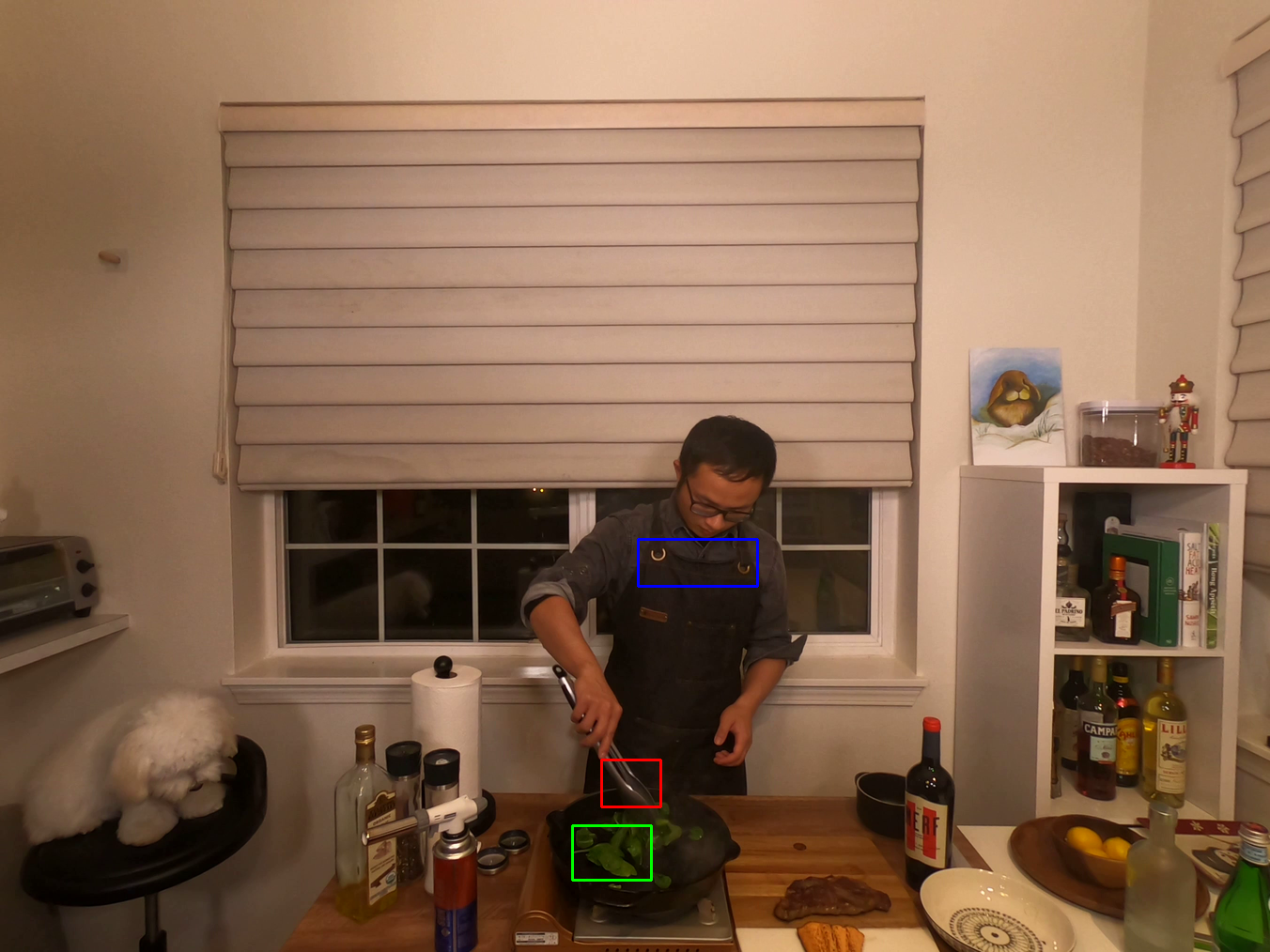}\\[-1pt]
    
    \begin{minipage}[b]{0.47\linewidth}
      \includegraphics[width=\linewidth]{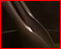}
    \end{minipage}%
    \begin{minipage}[b]{0.538\linewidth}
      \includegraphics[width=\linewidth]{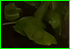}
    \end{minipage}\\[-1pt]
    
    \includegraphics[width=\linewidth]{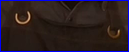}
    \caption*{GT}
    \end{subfigure}

  \caption{Visual comparisons on Neu3D dataset. The zoomed views reveal that the baselines produce over-smoothed structures on the metal strap (red box), the spinach leaves (green box), and the ring details on the clothes (blue box). Incorporating our temporal deformation evolution modeling substantially sharpens these structures, restores richer local contrast, and yields more faithful fine-scale geometry and appearance across both backbones.}
  \label{fig_neu3d}
\end{figure*}

\vspace{-3pt}
\noindent \textbf{Uncertainty-aware Trigger.}
In 3DGS-style pipelines, densification is typically decided solely by the gradient magnitude $\rho_i$, which may prematurely densify Gaussians under motion ambiguity. We augment the trigger with two uncertainty cues derived from Sec.~\ref{sec:deform_correction}: the prediction–observation residual $\bm{r}_{g_i}^t$ and the angular deviation (defined in Eq.~\ref{eq_proxy_angle}). Concretely, a Gaussian $\bm{g}_i$ is densified only if
\begin{equation}
\rho_i < \tau_\rho, \qquad r_i^t < \tau_r, \qquad \theta_i^{t} < \tau_\theta,
\label{eq:densi_trigger}
\end{equation}
where $\tau_\rho$ is the standard gradient threshold and $(\tau_r,\tau_\theta)$ are fixed across scenes to avoid overfitting. Small $r_i^t$ and $\theta_i^t$ indicate temporally consistent motion and a reliable prior, reducing the chance of propagating uncertainty.

\noindent \textbf{Deformation-aware Clone/Split.}
Standard densification operates at the source Gaussian’s current position: a \emph{clone} duplicates a small Gaussian with parameters identical to the source, and a \emph{split} replaces a large Gaussian with two smaller components (\textit{e.g.}, size scaled by $1/1.6$) whose positions are sampled from a Gaussian defined by the source’s mean and covariance. 
To better align with motion, we first apply a partial deformation to the source,
\begin{equation}
\bm{g}_i \leftarrow \bm{g}_i + \alpha\,\tilde{\bm{d}}_{g_i}^t, \qquad \alpha=0.5,
\label{eq:deformation-clone}
\end{equation}
and then perform clone/split at the updated location as shown in Fig.~\ref{fig_densify}. The resulting Gaussians inherit shape/opacity/color following standard rules.

\noindent\textbf{Buffer Maintenance during Gaussian Control.}
Since EvoGS maintains persistent deformation states for each Gaussian, the state buffers must be updated consistently when the Gaussian set changes during densification and pruning. When new Gaussians are instantiated by clone or split, we initialize their deformation buffers by inheriting the corrected deformation history from their parent Gaussians. Specifically, for a newly created Gaussian $\bm{g}_{new}$ from parent $\bm{g}_{p}$, we set
\begin{equation}
\mathcal{B}_{g_{new}} \leftarrow \mathcal{B}_{g_{p}},
\qquad
\bm{m}_{g_{new}} \leftarrow \bm{m}_{g_{p}},
\label{eq:buffer_inheritance}
\end{equation}
where $\mathcal{B}$ denotes the persistent deformation buffer and $\bm{m}$ denotes the innovation memory used for adaptive correction. This inheritance provides newly instantiated Gaussians with a valid local motion history, avoiding cold-start artifacts in subsequent state extrapolation. During pruning, the deformation buffers and innovation memories are filtered using the same validity mask as the Gaussian parameters, ensuring that the temporal states remain aligned with the active Gaussian set.

\begin{table*}[!h]
\centering
\setlength{\tabcolsep}{1.5pt}
\caption{Quantitative comparisons on Neu3D dataset. All methods are re-trained and metrics are evaluated at $1352 \times 1014$ resolution, complexity results are obtained using one H200 GPU. The \colorbox{\best}{best}, \colorbox{\second}{second best} and \colorbox{\third}{third best} performances are colored. EvoGS consistently outperforms all baselines by substantial margins, demonstrating the effectiveness of our temporal deformation evolution modeling. Furthermore, with the proposed deformation-aware Gaussian control strategy, EvoGS achieves improved efficiency, yielding faster rendering speed and shorter training time.}
\resizebox{0.86\linewidth}{!}{
\begin{tabular}{c| ccc| ccc| ccc| ccc}
\toprule

\multirow{2}{*}{Methods}&
\multicolumn{3}{c|}{\textit{coffee\_martini}}&
\multicolumn{3}{c|}{\textit{cook\_spinach}}&
\multicolumn{3}{c|}{\textit{cut\_roasted\_beef}}&
\multicolumn{3}{c}{\textit{flame\_salmon\_1}}\cr
\cmidrule(lr){2-4}\cmidrule(lr){5-7}\cmidrule(lr){8-10}\cmidrule(lr){11-13}
&PSNR$\uparrow$&SSIM$\uparrow$&LPIPS$\downarrow$&
PSNR$\uparrow$&SSIM$\uparrow$&LPIPS$\downarrow$&
PSNR$\uparrow$&SSIM$\uparrow$&LPIPS$\downarrow$&
PSNR$\uparrow$&SSIM$\uparrow$&LPIPS$\downarrow$\cr
\midrule

4DGS~\cite{4DGS}
&28.41 &0.908 &0.162
&\cellcolor{\best}32.22  &0.938 &0.170
&29.68  &0.932 &0.164
&29.27 &0.913   &0.155 \cr

\textbf{EvoGS(4DGS)}
&\cellcolor{\third}29.68 &0.915  &0.159
&\cellcolor{\second}33.11 &0.943 &0.158
&\cellcolor{\second}32.82 &\cellcolor{\second}0.949 &0.149
&\cellcolor{\third}30.14 &0.922  &0.146 \cr

Grid4D~\cite{grid4d}
&28.19 &0.899&0.175
&32.30  &0.945 &\cellcolor{\third}0.149
&29.79 &0.918 &0.190
&29.05 &0.910 &0.166 \cr

\textbf{EvoGS(Grid4D)}
&\cellcolor{\second}30.35 &\cellcolor{\second}0.919 &0.156
&32.81  &\cellcolor{\best}0.949 &\cellcolor{\best}0.147
&\cellcolor{\best}33.26 &\cellcolor{\best}0.950 &\cellcolor{\best}0.145
&\cellcolor{\second}30.60 &\cellcolor{\third}0.923 &0.150 \cr

Motion-GS~\cite{motiongs}
&27.14 &0.900 &0.156
&31.16  &0.943 &0.149
&30.74 &0.942 &0.151
&27.10 &0.907   &0.148 \cr

\textbf{EvoGS(Motion-GS)}
&28.56 &\cellcolor{\third}0.919 &\cellcolor{\third}0.149
&31.77 &\cellcolor{\second}0.947 &\cellcolor{\second}0.147
&\cellcolor{\third}31.96 &\cellcolor{\third}0.948 &\cellcolor{\third}0.146
&28.20& 0.919  & \cellcolor{\third}0.143 \cr

MoDec-GS~\cite{motiongs}
&28.34 &0.918 &\cellcolor{\second}0.142
&28.78 &0.938 &0.154
&26.42 &0.926 &0.156
&29.05 &\cellcolor{\second}0.924 &\cellcolor{\second}0.133 \cr

\textbf{EvoGS(MoDec-GS)}
&\cellcolor{\best}32.04 &\cellcolor{\best}0.941& \cellcolor{\best}0.131
&29.60  &\cellcolor{\third}0.946 &0.150
&29.57 &0.946 & \cellcolor{\second}0.145
&\cellcolor{\best}31.96 &\cellcolor{\best}0.946 &\cellcolor{\best}0.125 \cr

DASH~\cite{dash}
&26.95 &0.895&0.173
&27.65  &0.913 &0.193
&27.32 &0.918 &0.194
&27.06 &0.900 &0.171 \cr

\textbf{EvoGS(DASH)}
&27.79 &0.905   &0.169
&28.24 &0.914 &0.192
&27.93 &0.921 &0.180
&27.98 &0.909   &0.161 \cr

SpeeDe3DGS~\cite{tu2026speede3dgs}
&26.13 &0.887 &0.191
&29.77  &0.928 &0.189
&23.86 &0.905 &0.217
&27.12 &0.911   &0.152 \cr

\textbf{EvoGS(SpeeDe3DGS)}
&26.89 &0.898 &0.187
&30.54 &0.935 &0.181
&29.83 &0.928 &0.198
&28.37 &0.915 &0.169 \cr

\midrule

\multirow{2}{*}{Methods}&
\multicolumn{3}{c|}{\textit{flame\_steak}}&
\multicolumn{3}{c|}{\textit{sear\_steak}}&
\multicolumn{3}{c|}{\textit{Average}}&
\multicolumn{3}{c}{\textit{Complexity (coffee\_martini)}}\cr
\cmidrule(lr){2-4}\cmidrule(lr){5-7}\cmidrule(lr){8-10}\cmidrule(lr){11-13}
&PSNR$\uparrow$&SSIM$\uparrow$&LPIPS$\downarrow$&
PSNR$\uparrow$&SSIM$\uparrow$&LPIPS$\downarrow$&
PSNR$\uparrow$&SSIM$\uparrow$&LPIPS$\downarrow$&
Num of Gaussians$\downarrow$&FPS$\uparrow$&Training Time$\downarrow$\cr
\midrule

4DGS~\cite{4DGS}
&29.99 &0.941 &0.158
&32.52  &0.949 &0.153
&30.35 &0.930 &0.160
&125 K  &41 &\cellcolor{\second}24 min  \cr

\textbf{EvoGS(4DGS)}
&\cellcolor{\second}33.55 &\cellcolor{\third}0.953 &0.142
&\cellcolor{\best}33.84  &0.955 &0.140
&\cellcolor{\second}32.19 &\cellcolor{\third}0.940 &0.149
&122 K  &56 &\cellcolor{\best}13 min \cr

Grid4D~\cite{grid4d}
&\cellcolor{\third}32.32 &0.952 &0.139
&\cellcolor{\third}32.86  &\cellcolor{\third}0.957 &\cellcolor{\second}0.135
&\cellcolor{\third}30.75 &0.930 &0.159
&164 K &217   &49 min  \cr

\textbf{EvoGS(Grid4D)}
&\cellcolor{\best}33.83 &\cellcolor{\best}0.959 &\cellcolor{\best}0.132
&\cellcolor{\second}33.77  &\cellcolor{\best}0.961 &\cellcolor{\best}0.130
&\cellcolor{\best}32.44 &\cellcolor{\second}0.944&\cellcolor{\third}0.143
&151 K& \cellcolor{\second}244  & 37 min\cr

Motion-GS~\cite{motiongs}
&31.37 &0.952 &\cellcolor{\third}0.134
&28.29  &0.941 &0.142
&29.30 &0.931 &0.147
&728 K &26   &363 min\cr

\textbf{EvoGS(Motion-GS)}
&32.10 & \cellcolor{\second}0.956& \cellcolor{\second}0.132
&29.54  &0.947 &0.139
&30.36 &0.939 &\cellcolor{\second}0.143
&637 K &58   &287 min\cr

MoDec-GS~\cite{modecgs}
&24.21 &0.854 &0.313
&23.61  &0.907 &0.166
&26.74 &0.911 &0.177
&98 K &27  &41 min\cr

\textbf{EvoGS(MoDec-GS)}
&28.84 &0.953&0.144
&30.55 &\cellcolor{\second}0.958&\cellcolor{\third}0.138
&30.43 &\cellcolor{\best}0.948 &\cellcolor{\best}0.139
&\cellcolor{\third}91 K&45 &36 min\cr

DASH~\cite{dash}
&27.59 &0.924 &0.181
&29.98 &0.930 &0.174
&27.76 &0.913 &0.181
&206 K &159   &40 min \cr

\textbf{EvoGS(DASH)}
&28.15 &0.930 &0.174
&30.54 &0.936 &0.167
&28.44 &0.919 &0.174
&197 K &175 &\cellcolor{\third}35 min \cr

SpeeDe3DGS~\cite{tu2026speede3dgs}
&25.51 &0.919&0.194
&27.14  &0.922 &0.193
&26.59 &0.912 &0.189
&\cellcolor{\second}40 K&\cellcolor{\third}224  &44 min \cr

\textbf{EvoGS(SpeeDe3DGS)}
&29.96 &0.940 &0.174
&29.04  &0.933 &0.175
&29.11 &0.925 &0.181
&\cellcolor{\best}37 K & \cellcolor{\best}246  &41 min \cr

\bottomrule
\end{tabular}
}
\label{tab_neu3d_comparison}
\end{table*}

\begin{figure}[!h]
    \centering
    \includegraphics[width=0.96\linewidth]{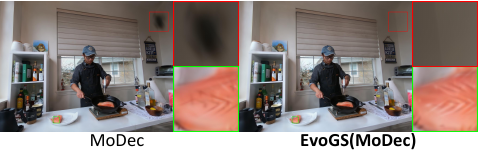}
    \caption{MoDec produces local blurred artifacts, while EvoGS yields cleaner reconstruction with sharper structural details.}
    \label{fig_neu2}
\end{figure}

\subsection{Optimization}
\label{sec_optimization}

To optimize Gaussians $\mathcal{G}$, the observation network $\bm \Phi$, and the gain estimator $\bm \Psi$, the render loss is adopted by comparing the rendered image at different time steps with ground truth reference images using a combination of $\mathcal{L}_1$ loss and D-SSIM loss following baseline methods.
Besides, specific in our method, our training objective integrates two auxiliary loss terms to enhance the deformation estimation process, targeting complementary aspects of estimation accuracy and predictive stability.

\noindent \textbf{Deformation supervision.}
To ensure that the final corrected deformation offset aligns with the local motion evidence encoded in the observation, we apply direct supervision by minimizing their discrepancy for all $N$ Gaussians:
\begin{equation}
\mathcal{L}_{\text{sup}} = \frac{1}{N} \sum_{i=1}^{N} \left\| \tilde{\bm{d}}_{g_i}^t - \bm{z}_{g_i}^t  \right\|_2^2.
\end{equation}
This term encourages the corrected deformation state to remain consistent with the observation-driven displacement, providing a per-frame corrective constraint for the adaptive deformation correction.
 
It is important when the prediction or prior motion is unreliable, allowing the network to compensate through observations.
%particularly

\noindent \textbf{Temporal prior regularization.}
The extrapolated prior $\hat{\bm{d}}_i^t$ is computed from the persistent deformation buffer and contains no additional learnable parameters. We therefore use it as a temporal reference to weakly regularize the current MLP-derived observation:
\begin{equation}
\mathcal{L}_{\text{prior}} =
\frac{1}{N}
\sum_{i=1}^{N}
sg(\sigma_i)
\left\|
sg(\hat{\bm{d}}_{g_i}^t) - \bm{z}_{g_i}^t
\right\|_2^2 ,
\end{equation}
where $sg(\cdot)$ denotes the stop-gradient operation. This term optimizes the observation network $\boldsymbol{\Phi}$ by discouraging abrupt deviations from the historical motion prior in visible regions, while the adaptive correction module still determines how much the final state should rely on the prior or the current observation.

Therefore, the final optimization objective is a weighted sum of the above components:
\begin{equation}
\mathcal{L}_{\text{total}} = \mathcal{L}_{\text{render}} + \lambda_1 \mathcal{L}_{\text{sup}} + \lambda_2 \mathcal{L}_{\text{prior}},
\end{equation}
where $\lambda_1$ and $\lambda_2$ are hyper-parameters, and we set $\lambda_1 =\lambda_2 = 0.5$ in our implementations. Notably, $\mathcal{L}_{\text{sup}}$ and $ \mathcal{L}_{\text{pred}}$ are calculated within the deformation network without the extra forward rendering or backpropagation over Gaussians, achieving superior performance with minimal overhead.

\begin {table*}[!t]
\centering
%\footnotesize
\setlength{\abovecaptionskip}{2mm}
\setlength\tabcolsep{0.8pt}
\renewcommand{\arraystretch}{1}
%\vspace{-1mm}
\caption{\textbf{Quantitative comparison on NeRF-DS dataset}. Our temporal deformation evolution modeling replaces the original deformation modules in various deformable Gaussian frameworks (\textit{e.g.}, 4DGS, SCGS, MAGS, D-3DGS), forming their EvoGS variants.
These EvoGS models consistently achieve higher PSNR/SSIM and lower LPIPS across diverse dynamic scenes, demonstrating the robustness and broad applicability of our temporal deformation evolution modeling.}
\resizebox{0.86\textwidth}{!}{
\begin{tabular}{ccccccccccccccccc}
\toprule[1pt]
\multirow{2}{*}{Methods} &
  \multicolumn{3}{c}{\textbf{Sieve}} &
  \multicolumn{3}{c}{\textbf{Plate}} &
  \multicolumn{3}{c}{\textbf{Bell }} &
  \multicolumn{3}{c}{\textbf{Press}} \\
\cmidrule(lr){2-4} \cmidrule(lr){5-7} \cmidrule(lr){8-10} \cmidrule(lr){11-13}
& PSNR($\uparrow$) & SSIM($\uparrow$) & LPIPS($\downarrow$)
& PSNR($\uparrow$) & SSIM($\uparrow$) & LPIPS($\downarrow$)
& PSNR($\uparrow$) & SSIM($\uparrow$) & LPIPS($\downarrow$)
& PSNR($\uparrow$) & SSIM($\uparrow$) & LPIPS($\downarrow$) \\
\midrule

3D-GS~\cite{3Dgaussian} & 23.16 & 0.8203 & 0.2247 & 16.14 & 0.6970 & 0.4093 & 21.01 & 0.7885 & 0.2503 & 22.89 & 0.8163 & 0.2904 \\

TiNeuVox~\cite{TiNeuVox} & 21.49 & 0.8265 & 0.3176 & \cellcolor{\third}20.58 & 0.8027 & 0.3317 & 23.08 & 0.8242 & 0.2568 & 24.47 & 0.8613 & 0.3001 \\

HyperNeRF~\cite{park2021hypernerf} & 25.43 & 0.8798 & 0.1645 & 18.93 & 0.7709 & 0.2940 & 23.06 & 0.8097 & 0.2052 & 26.15 & \cellcolor{\best}0.8897 & 0.1959 \\

NeRF-DS~\cite{nerfds} & 25.78 & \cellcolor{\second}0.8900 & \cellcolor{\second}0.1472 & 20.54 & 0.8042 & 0.1996 & 23.19 & 0.8212 & 0.1867 & 25.72 & 0.8618 & 0.2047 \\
\midrule

4DGS~\cite{4DGS} & 24.33 & 0.8625 & 0.1583 & 18.07 & 0.7215 & 0.3218 & 24.44 & 0.8237 & 0.1725 & 24.57 & 0.8611 & 0.2039 \\

\textbf{EvoGS(4DGS)} & 25.74 & 0.8827 & 0.1484 & 20.33 & 0.7968 & 0.2009 & 25.96 & 0.8525 & 0.1530 & 26.03 & 0.8699 & 0.1836 \\

\midrule

SCGS~\cite{SC-GS} & 24.93 & 0.8622 & 0.1563 & 19.08 & 0.7488 & 0.3620 & 25.11 & 0.8425 & 0.1628 & 25.40 & 0.8609 & 0.1943 \\

\textbf{EvoGS(SCGS)} & \cellcolor{\second}26.25 & \cellcolor{\third}0.8875 & \cellcolor{\third}0.1474 & 20.55 & \cellcolor{\third}0.8127 & \cellcolor{\third}0.1901 & \cellcolor{\second}26.44 & \cellcolor{\second}0.8674 & \cellcolor{\second}0.1451 & \cellcolor{\second}26.33 & 0.8681 & \cellcolor{\third}0.1827 \\

\midrule

MAGS~\cite{mags} & 24.82 & 0.8673 & 0.1533 & 18.83 & 0.7413 & 0.3486 & 24.92 & 0.8417 & 0.1641 & 25.02 & 0.8644 & 0.1985 \\

\textbf{EvoGS(MAGS)} & \cellcolor{\third}26.23 & 0.8866 & 0.1474 & \cellcolor{\second}20.99 & \cellcolor{\second}0.8211 & \cellcolor{\second}0.1872 & \cellcolor{\third}26.35 & \cellcolor{\third}0.8652 & \cellcolor{\third}0.1481 & \cellcolor{\third}26.28 & \cellcolor{\third}0.8712 & \cellcolor{\second}0.1816 \\

\midrule

D-3DGS~\cite{deformablegs} & 25.24 & 0.8686 & 0.1505 & 19.23 & 0.7523 & 0.3576 & 25.38 & 0.8473 & 0.1593 & 25.40 & 0.8609 & 0.1943 \\

\textbf{EvoGS(D-3DGS)} & \cellcolor{\best}26.46 & \cellcolor{\best}0.8916 & \cellcolor{\best}0.1454 & \cellcolor{\best}21.28 & \cellcolor{\best}0.8241 & \cellcolor{\best}0.1831 & \cellcolor{\best}26.82 & \cellcolor{\best}0.8705 & \cellcolor{\best}0.1419 & \cellcolor{\best}26.81 & \cellcolor{\second}0.8745 & \cellcolor{\best}0.1789 \\

\midrule[1pt]  % 粗线分隔上半部分与下半部分
\multirow{2}{*}{Methods} &
  \multicolumn{3}{c}{\textbf{Cup}} &
  \multicolumn{3}{c}{\textbf{As}} &
  \multicolumn{3}{c}{\textbf{Basin}} &
  \multicolumn{3}{c}{\textbf{\textit{Average}}} \\
\cmidrule(lr){2-4} \cmidrule(lr){5-7} \cmidrule(lr){8-10} \cmidrule(lr){11-13}
& PSNR($\uparrow$) & SSIM($\uparrow$) & LPIPS($\downarrow$)
& PSNR($\uparrow$) & SSIM($\uparrow$) & LPIPS($\downarrow$)
& PSNR($\uparrow$) & SSIM($\uparrow$) & LPIPS($\downarrow$)
& PSNR($\uparrow$) & SSIM($\uparrow$) & LPIPS($\downarrow$) \\
\midrule
3D-GS~\cite{3Dgaussian} & 21.71 & 0.8304 & 0.2548 & 22.69 & 0.8017 & 0.2994 & 18.42 & 0.7170 & 0.3153 & 20.86 & 0.7816 & 0.2920 \\

TiNeuVox~\cite{TiNeuVox} & 19.71 & 0.8109 & 0.3643 & 21.26 & 0.8289 & 0.3967 & \cellcolor{\second}20.66 & 0.8145 & 0.2690 & 21.61 & 0.8241 & 0.3195 \\

HyperNeRF~\cite{park2021hypernerf} & 24.59 & 0.8770 & 0.1650 & 25.58 & \cellcolor{\best}0.8949 & 0.1777 & 20.41 & 0.8199 & 0.1911 & 23.45 & 0.8488 & 0.1991 \\

NeRF-DS~\cite{nerfds} & 24.91 & 0.8741 & 0.1737 & 25.13 & 0.8778 & \cellcolor{\third}0.1741 & 19.96 & 0.8166 & 0.1855 & 23.60 & 0.8494 & 0.1816 \\

\midrule

4DGS~\cite{4DGS} & 23.85 & 0.8802 & 0.1645 & 25.26 & 0.8732 & 0.1942 & 18.66 & 0.7885 & 0.2048 & 22.74 & 0.8301 & 0.2029 \\

\textbf{EvoGS(4DGS)} & 24.83 & 0.8887 & 0.1542 & 26.33 & 0.8868 & 0.1799 & 19.99 & 0.7994 & 0.1876 & 24.17 & 0.8538 & 0.1725 \\

\midrule

SCGS~\cite{SC-GS} & 24.44 & 0.8875 & 0.1579 & 25.75 & 0.8796 & 0.1877 & 19.24 & 0.7902 & 0.1911 & 23.42 & 0.8388 & 0.2017 \\

\textbf{EvoGS(SCGS)} & \cellcolor{\second}25.56 & \cellcolor{\second}0.8934 & \cellcolor{\second}0.1501 & \cellcolor{\second}26.80 & \cellcolor{\third}0.8911 & \cellcolor{\second}0.1735 & \cellcolor{\third}20.65 & \cellcolor{\second}0.8218 & \cellcolor{\second}0.1819 & \cellcolor{\second}24.65 & \cellcolor{\third}0.8631 & \cellcolor{\second}0.1673 \\

\midrule

MAGS~\cite{mags} & 24.35 & 0.8824 & 0.1608 & 25.58 & 0.8756 & 0.1906 & 19.17 & 0.7868 & 0.1938 & 23.24 & 0.8371 & 0.2014 \\

\textbf{EvoGS(MAGS)} & \cellcolor{\third}25.28 & \cellcolor{\third}0.8923 & \cellcolor{\third}0.1524 & \cellcolor{\third}26.53 & 0.8886 & 0.1759 & 20.61 & \cellcolor{\third}0.8212 & \cellcolor{\third}0.1821 & \cellcolor{\third}24.61 & \cellcolor{\second}0.8637 & \cellcolor{\third}0.1678 \\

\midrule

D-3DGS~\cite{deformablegs} & 24.82 & 0.8909 & 0.1542 & 26.01 & 0.8822 & 0.1837 & 19.69 & 0.7935 & 0.1886 & 23.68 & 0.8422 & 0.1983 \\

\textbf{EvoGS(D-3DGS)} & \cellcolor{\best}25.89 & \cellcolor{\best}0.8951 & \cellcolor{\best}0.1472 & \cellcolor{\best}27.06 & \cellcolor{\second}0.8936 & \cellcolor{\best}0.1704 & \cellcolor{\best}20.98 & \cellcolor{\best}0.8251 & \cellcolor{\best}0.1755 & \cellcolor{\best}25.04 & \cellcolor{\best}0.8678 & \cellcolor{\best}0.1632 \\

\hline
\end{tabular}}

\label{tab_nerf-ds}

\end{table*}

\begin{figure*}[!h]
  \centering

  % ==== 第 1 行：4 列 ====
  \begin{subfigure}{0.245\linewidth}
    \centering
    \includegraphics[width=\linewidth]{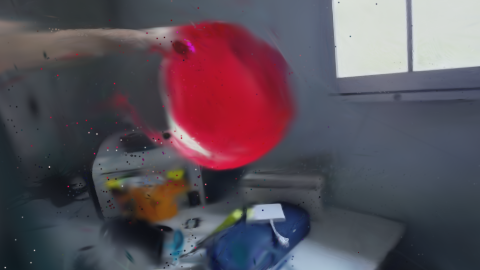}%
    \caption*{4DGS}
  \end{subfigure}
  \hfill
  \begin{subfigure}{0.245\linewidth}
    \centering
    \includegraphics[width=\linewidth]{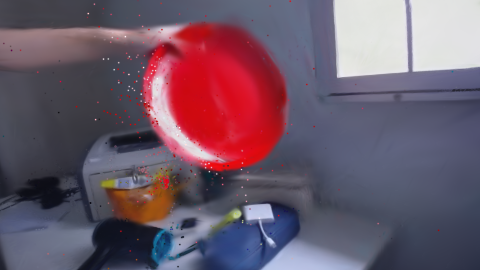}%
    \caption*{\textbf{EvoGS(4DGS)}}
  \end{subfigure}
  \hfill
  \begin{subfigure}{0.245\linewidth}
    \centering
    \includegraphics[width=\linewidth]{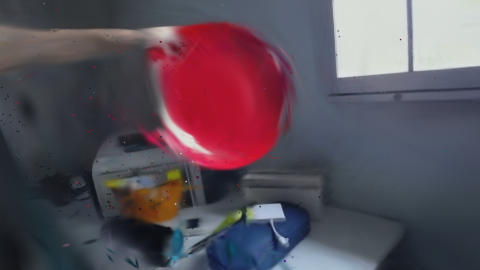}%
    \caption*{MAGS}
  \end{subfigure}
  \hfill
  \begin{subfigure}{0.245\linewidth}
    \centering
    \includegraphics[width=\linewidth]{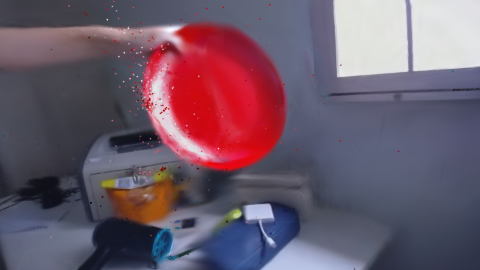}%
    \caption*{\textbf{EvoGS(MAGS)}}
  \end{subfigure}

  \vspace{2pt} 
  
  \begin{subfigure}{0.245\linewidth}
    \centering
    \includegraphics[width=\linewidth]{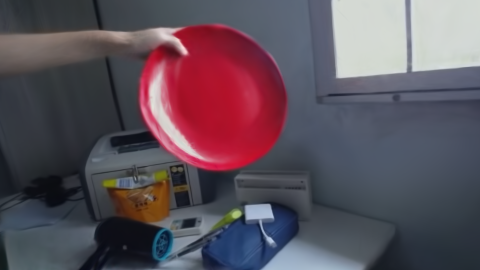}%
    \caption*{SCGS}
  \end{subfigure}
  \hfill
  \begin{subfigure}{0.245\linewidth}
    \centering
    \includegraphics[width=\linewidth]{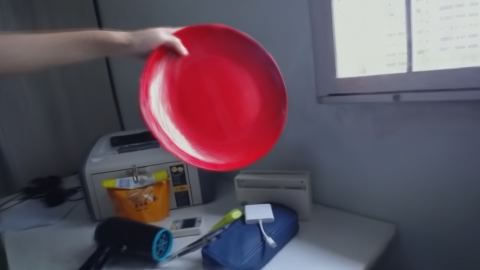}%
    \caption*{\textbf{EvoGS(SCGS)}}
  \end{subfigure}
  \hfill
  \begin{subfigure}{0.245\linewidth}
    \centering
    \includegraphics[width=\linewidth]{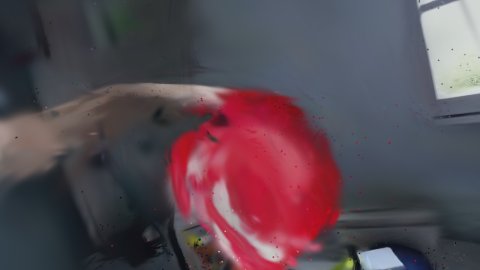}%
    \caption*{4DGS}
  \end{subfigure}
  \hfill
  \begin{subfigure}{0.245\linewidth}
    \centering
    \includegraphics[width=\linewidth]{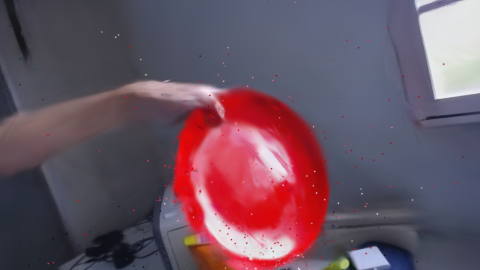}%
    \caption*{\textbf{EvoGS(4DGS)}}
  \end{subfigure}

  \vspace{2pt} 
  
  \begin{subfigure}{0.245\linewidth}
    \centering
    \includegraphics[width=\linewidth]{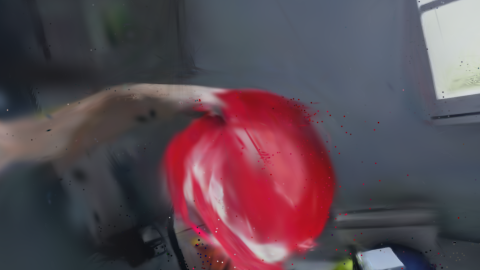}%
    \caption*{MAGS}
  \end{subfigure}
  \hfill
  \begin{subfigure}{0.245\linewidth}
    \centering
    \includegraphics[width=\linewidth]{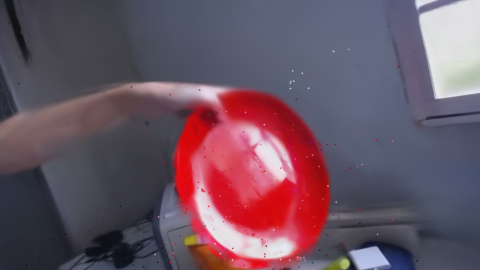}%
    \caption*{\textbf{EvoGS(MAGS)}}
  \end{subfigure}
  \hfill
  \begin{subfigure}{0.245\linewidth}
    \centering
    \includegraphics[width=\linewidth]{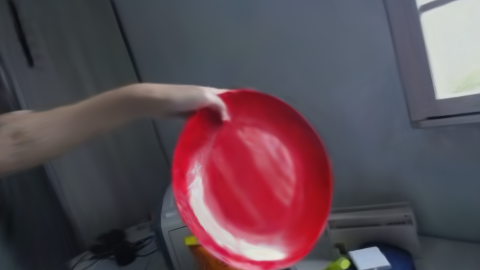}%
    \caption*{SCGS}
  \end{subfigure}
  \hfill
  \begin{subfigure}{0.245\linewidth}
    \centering
    \includegraphics[width=\linewidth]{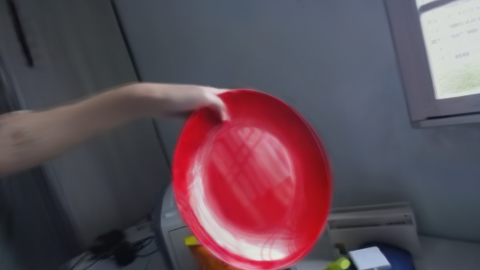}%
    \caption*{\textbf{EvoGS(SCGS)}}
  \end{subfigure}

  \caption{Qualitative comparison on the NeRF-DS dataset. The baselines (4DGS and MAGS) produce severely distorted object contours,  where the plate boundary and arm silhouette are noticeably warped. The EvoGS-augmented variants (\textbf{EvoGS(4DGS) and \textbf{EvoGS(MAGS)}}) preserve much more stable geometry and sharper outlines, demonstrating that the proposed temporal deformation evolution modeling consistently stabilizes geometry and appearance across diverse dynamic Gaussian frameworks.}
  \label{fig_visual_nerf-ds}
\end{figure*}

\begin{figure*}[t]
    \centering
    
    \newlength{\frameheight}
    \setlength{\frameheight}{5.4cm} 
    \setlength{\abovecaptionskip}{0mm}

    %=========== Row 1: 4DGS, 4DGS-KF, D-3DGS ===========
    % ---- 4DGS ----
    \begin{subfigure}{0.245\linewidth}
        \centering
        
        \begin{minipage}[b]{0.66\linewidth}
            \includegraphics[height=\frameheight]{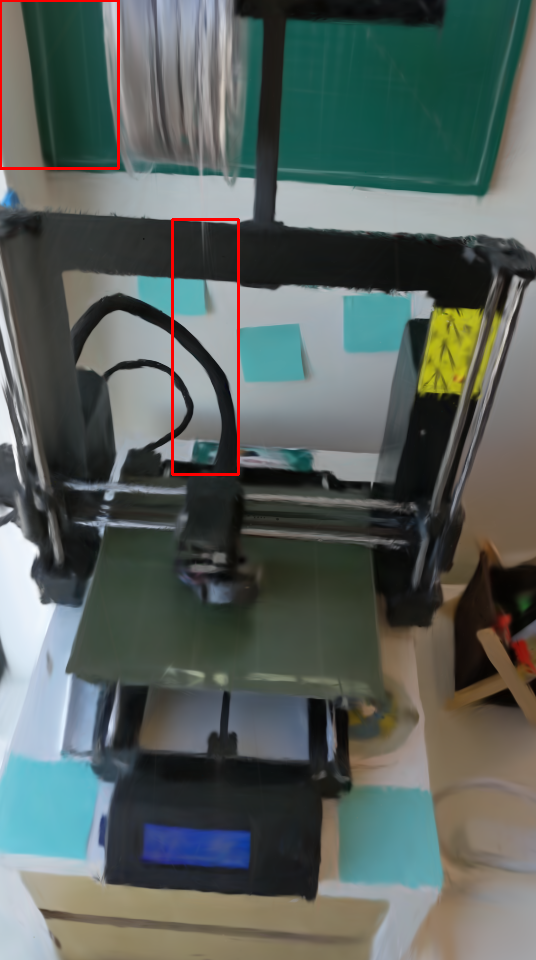}
        \end{minipage}%
        \hfill
        \begin{minipage}[b]{0.3\linewidth}
            \includegraphics[width=\linewidth,height=0.4\frameheight,keepaspectratio=false]{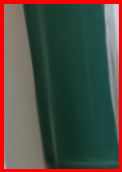}  \\[-1pt]
            \includegraphics[width=\linewidth,height=0.6\frameheight,keepaspectratio=false]{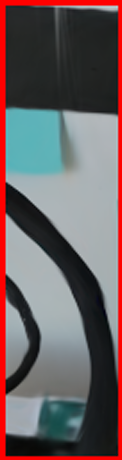}
        \end{minipage}
        \caption*{4DGS}
    \end{subfigure}
    \hfill
% --- 4DGS-KF ---
    \begin{subfigure}{0.245\linewidth}
        \centering
        \begin{minipage}[b]{0.66\linewidth}
            \includegraphics[height=\frameheight]{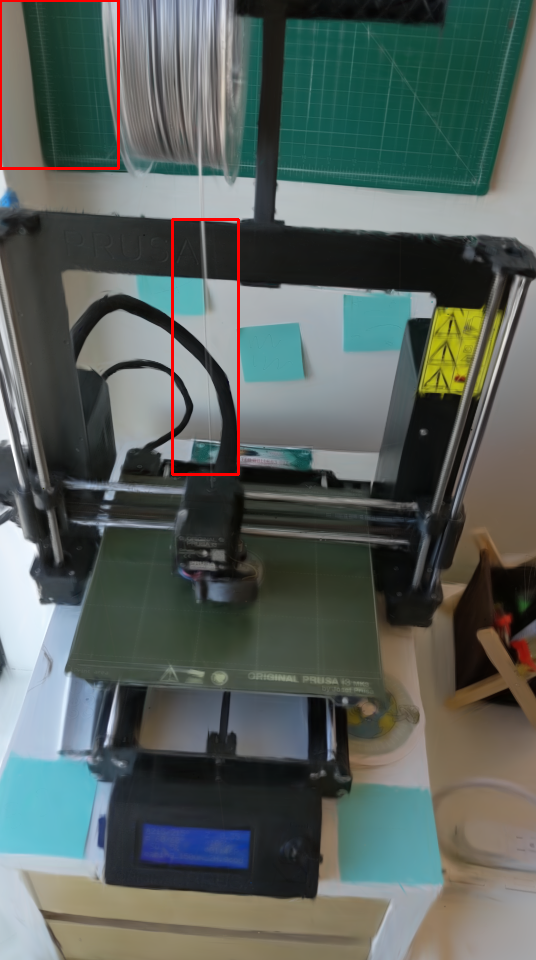}
        \end{minipage}%
        \hfill
        \begin{minipage}[b]{0.3\linewidth}
            \includegraphics[width=\linewidth,height=0.4\frameheight,keepaspectratio=false]{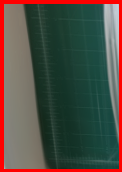}\\[-1pt]
            \includegraphics[width=\linewidth,height=0.6\frameheight,keepaspectratio=false]{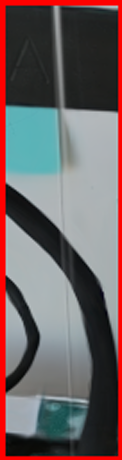}
        \end{minipage}
        \caption*{\textbf{EvoGS(4DGS)}}
    \end{subfigure}
    \hfill
    % ---- D-3DGS ----
    \begin{subfigure}{0.245\linewidth}
        \centering
        \begin{minipage}[b]{0.66\linewidth}
            \includegraphics[height=\frameheight]{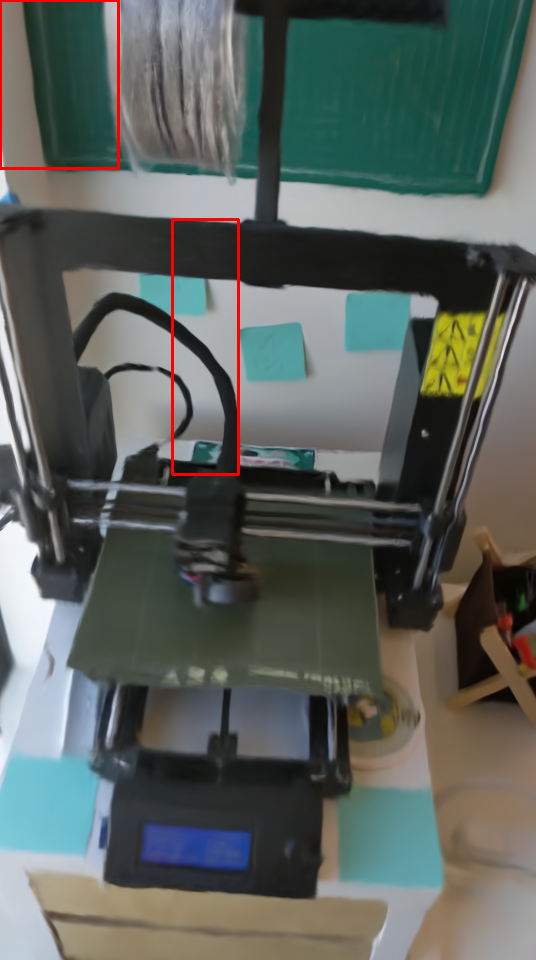}
        \end{minipage}%
        \hfill
        \begin{minipage}[b]{0.3\linewidth}
            \includegraphics[width=\linewidth,height=0.4\frameheight,keepaspectratio=false]{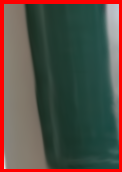}\\[-1pt]
            \includegraphics[width=\linewidth,height=0.6\frameheight,keepaspectratio=false]{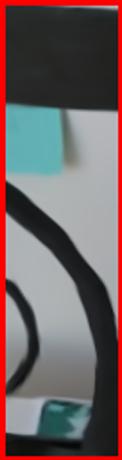}
        \end{minipage}
        \caption*{D-3DGS}
    \end{subfigure}
    \hfill
    \begin{subfigure}{0.245\linewidth}
        \centering
        \begin{minipage}[b]{0.66\linewidth}
            \includegraphics[height=\frameheight]{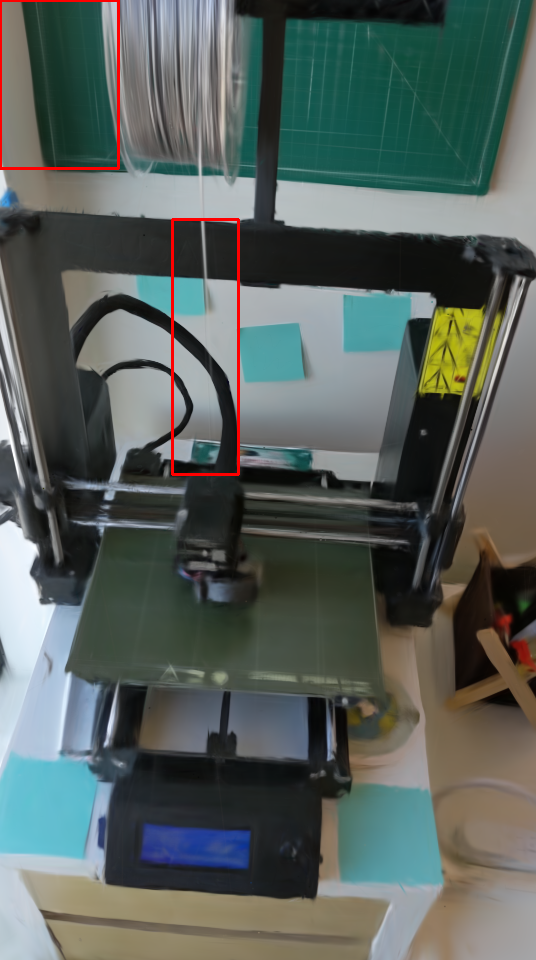}
        \end{minipage}%
        \hfill
        \begin{minipage}[b]{0.3\linewidth}
            \includegraphics[width=0.8\linewidth,height=0.4\frameheight,keepaspectratio=false]{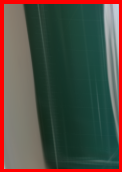}\\[-1pt]
            \includegraphics[width=0.8\linewidth,height=0.6\frameheight,keepaspectratio=false]{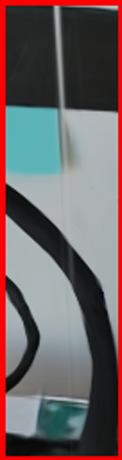}
        \end{minipage}
        \caption*{\textbf{EvoGS(D-3DGS)}}
    \end{subfigure}
    
    % ---- Grid4D ----
    \begin{subfigure}{0.245\linewidth}
        \centering
        \begin{minipage}[b]{0.66\linewidth}
            \includegraphics[height=\frameheight]{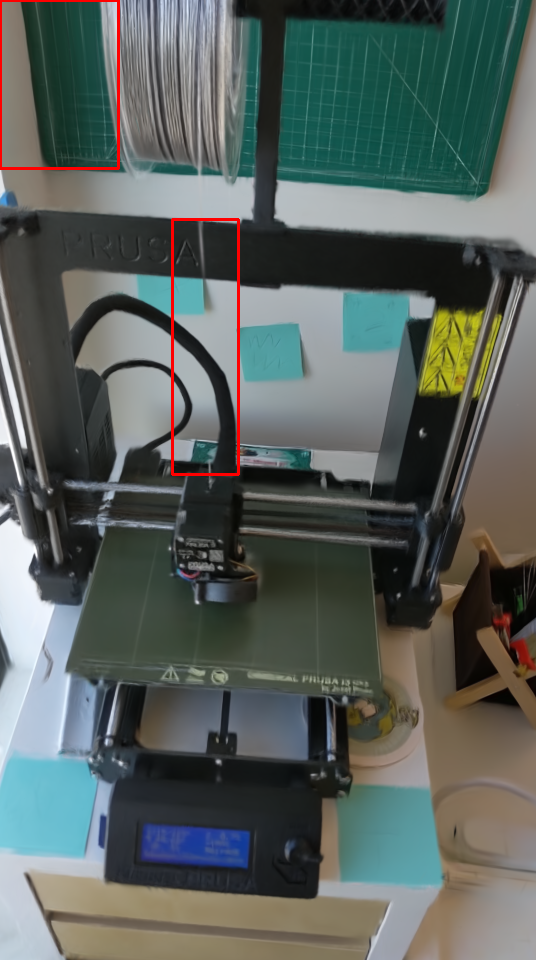}
        \end{minipage}%
        \hfill
        \begin{minipage}[b]{0.3\linewidth}
            \includegraphics[width=\linewidth,height=0.4\frameheight,keepaspectratio=false]{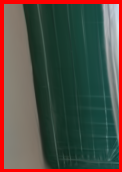}\\[-1pt]
            \includegraphics[width=\linewidth,height=0.6\frameheight,keepaspectratio=false]{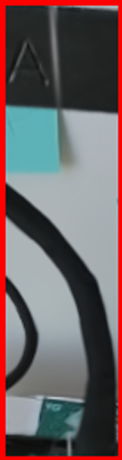}
        \end{minipage}
        \caption*{Grid4D}
    \end{subfigure}
    \hfill
    \begin{subfigure}{0.245\linewidth}
        \centering
        \begin{minipage}[b]{0.66\linewidth}
            \includegraphics[height=\frameheight]{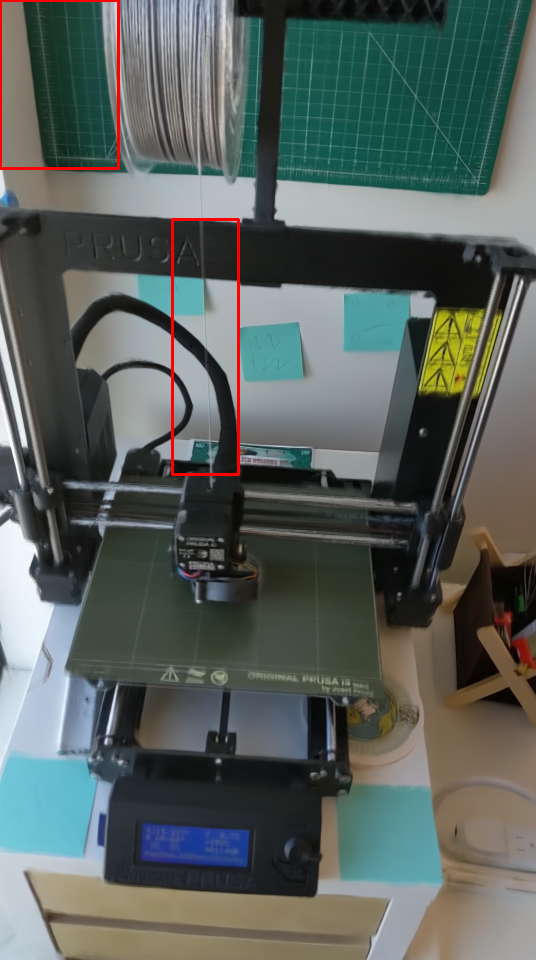}
        \end{minipage}%
        \hfill
        \begin{minipage}[b]{0.3\linewidth}
            \includegraphics[width=\linewidth,height=0.4\frameheight,keepaspectratio=false]{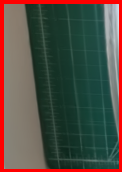}\\[-1pt]
            \includegraphics[width=\linewidth,height=0.6\frameheight,keepaspectratio=false]{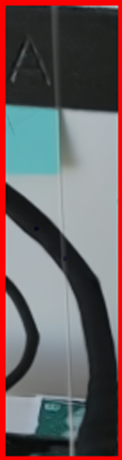}
        \end{minipage}
        \caption*{\textbf{EvoGS(Grid4D)}}
    \end{subfigure}
    \hfill
    % ---- MoDecGS ----
    \begin{subfigure}{0.245\linewidth}
        \centering
        \begin{minipage}[b]{0.66\linewidth}
            \includegraphics[height=\frameheight]{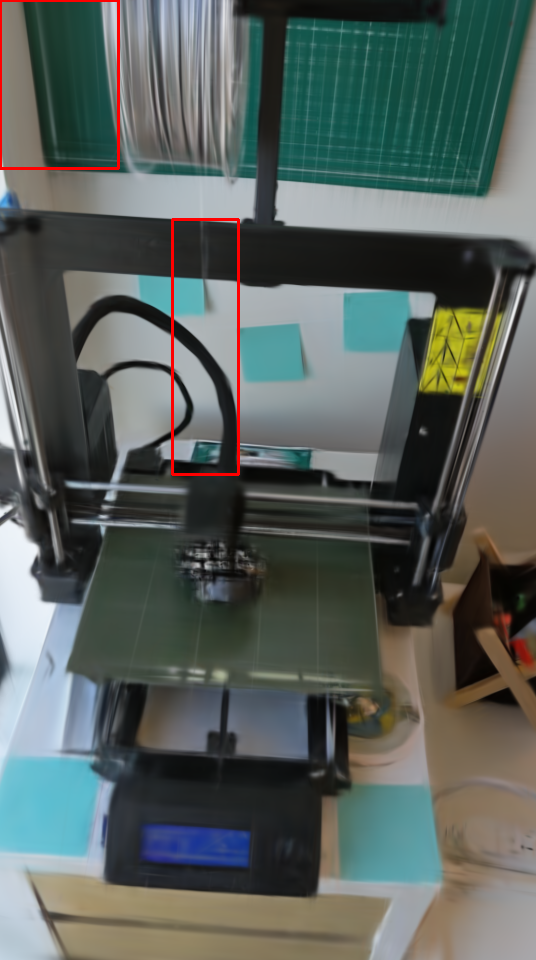}
        \end{minipage}%
        \hfill
        \begin{minipage}[b]{0.3\linewidth}
            \includegraphics[width=\linewidth,height=0.4\frameheight,keepaspectratio=false]{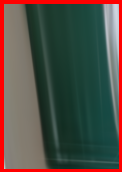}\\[-1pt]
            \includegraphics[width=\linewidth,height=0.6\frameheight,keepaspectratio=false]{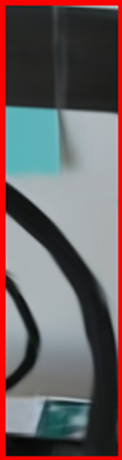}
        \end{minipage}
        \caption*{MoDecGS}
    \end{subfigure}
    \hfill
    % ---- MoDecGS-KF ----
    \begin{subfigure}{0.245\linewidth}
        \centering
        \begin{minipage}[b]{0.66\linewidth}
            \includegraphics[height=\frameheight]{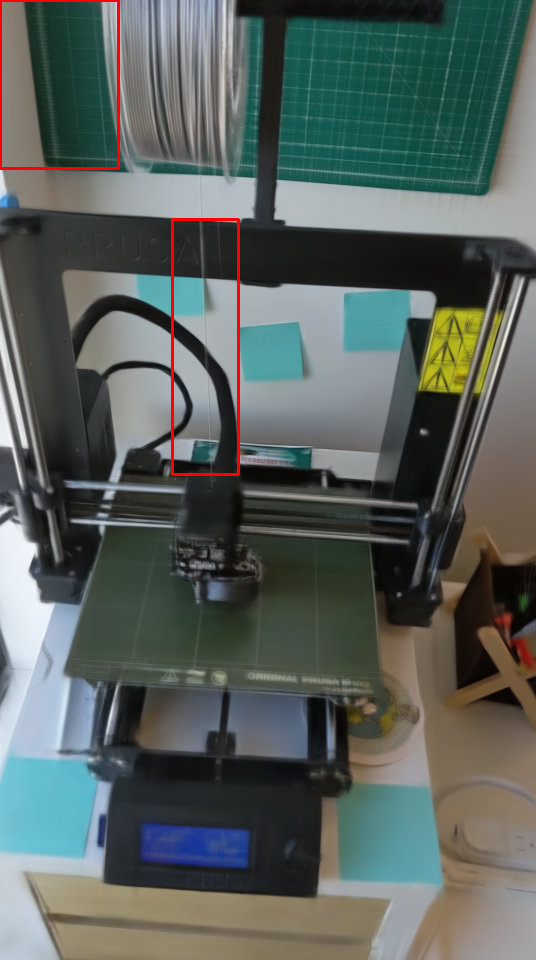}
        \end{minipage}%
        \hfill
        \begin{minipage}[b]{0.3\linewidth}
            \includegraphics[width=\linewidth,height=0.4\frameheight,keepaspectratio=false]{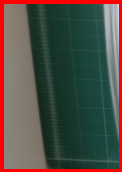}\\[-1pt]
            \includegraphics[width=\linewidth,height=0.6\frameheight,keepaspectratio=false]{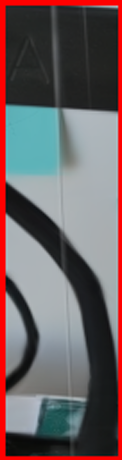}
        \end{minipage}
        \caption*{\textbf{EvoGS(MoDecGS)}}
    \end{subfigure}

 \caption{Qualitative comparisons on a complex scene with thin structures and sharp planar geometry (\textit{3D printer} scene in HyperNeRF dataset). The zoomed crops highlight that our EvoGS variants built upon various baselines enhance boundary localization and structural fidelity: vertical edges on the cutting mat become straighter and less wavy, the black cable and metallic frame exhibit fewer double edges and ringing artifacts, and small high-contrast details are better preserved, demonstrating consistent improvements across compared baselines.}
    \label{fig_visual_hypernerf}
\end{figure*}

\section{Experiments}
\label{sec_expriments}
\subsection{Experiment Setup} 
\noindent \textbf{Datasets and Metrics.} To assess the effectiveness of our method, we evaluate EvoGS on several datasets (NeRF-DS~\cite{nerfds}, HyperNeRF (vrig subset)~\cite{park2021hypernerf}, and Neu3D~\cite{neu3D}). NeRF-DS consists of seven real-world video sequences, with camera trajectories reconstructed using COLMAP~\cite{colmap}. Following the protocol in~\cite{4DGS}, we estimate camera poses and initialize Gaussians for HyperNeRF and Neu3D datasets. Training is conducted on the official training set, with evaluation carried out on the corresponding test scenes. We adopt the Peak Signal-to-Noise Ratio (PSNR) and Structural Similarity (SSIM~\cite{SSIM}) to assess pixel-level accuracy, and use Learned Perceptual Image Patch Similarity (LPIPS~\cite{LPIPS}, VGG) for perceptual quality evaluation.

\noindent \textbf{Implementation Details.} We conduct all the experiments on one H200 GPU device with Pytorch~\cite{pytorch} framework. {The inputs to the gain estimator are directly concatenated, including the temporal residual memory, the two deformation-velocity magnitudes, and the angular deviation, and then fed into the three-layer MLP. The output is a scalar correction weight for each Gaussian, which is shared across its deformation components. For the first two timestamps, where insufficient historical states are available for second-order extrapolation, we use the current observation to initialize the predicted state. At the third timestamp, second-order extrapolation is enabled with a fixed correction weight K=0.5. The learned adaptive correction is activated from the fourth timestamp, when sufficient historical states are available to compute the velocity and angular-deviation statistics. No additional warm-up stage is used. In the process of Gaussian densification, newly created Gaussians directly inherit the persistent deformation buffer and residual memory of their corresponding parent Gaussians for cloning; for splitting, the parent states are replicated for all generated child Gaussians, while their spatial parameters are initialized according to the standard split operation. We build our EvoGS by integrating our modular temporal deformation evolution process into several existing SOTA baselines (D-3DGS~\cite{deformablegs}, 4DGS~\cite{4DGS}, SCGS~\cite{SC-GS}, MAGS~\cite{mags}, Grid4D~\cite{grid4d}, MoDec-GS~\cite{modecgs}, DASH~\cite{dash}, SpeeDe3DGS~\cite{tu2026speede3dgs}). To ensure a fair comparison, the same deformation-aware densification strategy is applied to all integrated baselines and our training configurations (\textit{e.g.}, total training iterations and learning rate schedules) strictly follows the baselines. Furthermore, we replace the original deformation MLP/module in baselines with our observation network and temporal evolution/correction module and employ reduced channel dimensions for $\bm{\Phi}$ and $\bm{\Psi}$ (half of that of the original deformation module), thereby constraining their model capacity to be lower than that of the original deformation MLPs used in these baselines. 

\begin{table*}[!t]
\centering
\setlength{\tabcolsep}{1.5pt}
\setlength{\abovecaptionskip}{2.3mm}
\caption{Quantitative comparisons on the vrig subset of HyperNeRF, reported per scene. Metrics are evaluated at $536 \times 960$ resolution. Our Kalman-inspired deformation modeling can be seamlessly plugged into existing deformable Gaussian frameworks (\textit{e.g.}, 4DGS, D-3DGS, Grid4D, and MoDec-GS). The resulting variants (\textbf{EvoGS(4DGS), EvoGS(D-3DGS), EvoGS(Grid4D), EvoGS(MoDec-GS)}) consistently surpass their baselines in both PSNR and SSIM, highlighting the adaptability and generalization capability of our deformation design.}
\vspace{-1mm}
\resizebox{0.86\linewidth}{!}{
\begin{tabular}{c| cc| cc| cc| cc| cc}
\toprule
\multirow{2}{*}{Method}&
\multicolumn{2}{c}{\textbf{3D Printer}}&
\multicolumn{2}{c}{\textbf{Chicken}}&
\multicolumn{2}{c}{\textbf{Broom}}&
\multicolumn{2}{c}{\textbf{Banana}}&
\multicolumn{2}{c}{\textbf{Average}}\cr
\cmidrule(lr){2-3}\cmidrule(lr){4-5}\cmidrule(lr){6-7}\cmidrule(lr){8-9}\cmidrule(lr){10-11}
&PSNR$\uparrow$&SSIM$\uparrow$&
PSNR$\uparrow$&SSIM$\uparrow$&
PSNR$\uparrow$&SSIM$\uparrow$&
PSNR$\uparrow$&SSIM$\uparrow$&
PSNR$\uparrow$&SSIM$\uparrow$\cr

\midrule

HyperNeRF~\cite{park2021hypernerf}
&20.04&0.6308
&27.43&0.6312
&19.52&0.2114
&22.13&0.7209
&22.28&0.5486\cr

TiNeuVox~\cite{TiNeuVox}
&22.84&0.7316
&28.24&0.7923
&21.37&0.3138
&24.46&0.6419
&24.23&0.6199\cr

MotionGS~\cite{motiongs}
&20.63&0.6611
&22.87&0.6817
&21.04&0.3026
&26.57&0.8408
&22.78&0.6216\cr

\midrule

D-3DGS~\cite{deformablegs}
&20.56&0.6413
&22.84&0.6121
&20.58&0.3517
&26.04&0.8314
&22.50&0.6091\cr

\textbf{EvoGS(D-3DGS)}
&21.63\tiny\textcolor{red}{(+1.07)}
&0.6903\tiny\textcolor{red}{(+0.0490)}
&24.39\tiny\textcolor{red}{(+1.55)}
&0.7305\tiny\textcolor{red}{(+0.1184)}
&21.89\tiny\textcolor{red}{(+1.31)}
&0.3904\tiny\textcolor{red}{(+0.0387)}
&27.46\tiny\textcolor{red}{(+1.42)}
&0.8702\tiny\textcolor{red}{(+0.0388)}
&23.84\tiny\textcolor{red}{(+1.34)}
&0.6704\tiny\textcolor{red}{(+0.0613)}\cr

\midrule

4DGS~\cite{4DGS}
&22.05&0.7118
&28.76&0.8115
&22.14&0.3713
&28.07&0.8617
&25.26&0.6891\cr

\textbf{EvoGS(4DGS)}
&\cellcolor{\third}23.06\tiny\textcolor{red}{(+1.01)}
&\cellcolor{\third}0.7504\tiny\textcolor{red}{(+0.0386)}
&\cellcolor{\third}29.64\tiny\textcolor{red}{(+0.88)}
&\cellcolor{\third}0.8504\tiny\textcolor{red}{(+0.0389)}
&\cellcolor{\second}22.89\tiny\textcolor{red}{(+0.75)}
&0.4003\tiny\textcolor{red}{(+0.0290)}
&\cellcolor{\best}28.92\tiny\textcolor{red}{(+0.85)}
&\cellcolor{\third}0.9002\tiny\textcolor{red}{(+0.0385)}
&\cellcolor{\second}26.13\tiny\textcolor{red}{(+0.87)}
&\cellcolor{\third}0.7253\tiny\textcolor{red}{(+0.0362)}\cr

\midrule

Grid4D~\cite{grid4d}
&22.43&0.7235
&29.14&0.8431
&22.33&0.3739
&27.92&0.8625
&25.45&0.7007\cr

\textbf{EvoGS(Grid4D)}
&\cellcolor{\best}23.85\tiny\textcolor{red}{(+1.42)}
&\cellcolor{\best}0.7937\tiny\textcolor{red}{(+0.0702)}
&\cellcolor{\best}30.27\tiny\textcolor{red}{(+1.13)}
&\cellcolor{\best}0.8826\tiny\textcolor{red}{(+0.0395)}
&\cellcolor{\best}23.29\tiny\textcolor{red}{(+0.96)}
&\cellcolor{\third}0.4007\tiny\textcolor{red}{(+0.0268)}
&\cellcolor{\second}28.85\tiny\textcolor{red}{(+0.93)}
&\cellcolor{\best}0.9074\tiny\textcolor{red}{(+0.0449)}
&\cellcolor{\best}26.57\tiny\textcolor{red}{(+1.11)}
&\cellcolor{\best}0.7461\tiny\textcolor{red}{(+0.0454)}\cr

\midrule

MoDec-GS~\cite{modecgs}
&22.03&0.7023
&28.24&0.8116
&21.62&\cellcolor{\second}0.4033
&26.24&0.8409
&24.53&0.6895\cr

\textbf{EvoGS(MoDec-GS)}
&\cellcolor{\second}23.36\tiny\textcolor{red}{(+1.33)}
&\cellcolor{\second}0.7628\tiny\textcolor{red}{(+0.0605)}
&\cellcolor{\second}29.66\tiny\textcolor{red}{(+1.42)}
&\cellcolor{\second}0.8653\tiny\textcolor{red}{(+0.0537)}
&\cellcolor{\third}22.79\tiny\textcolor{red}{(+1.17)}
&\cellcolor{\best}0.4284\tiny\textcolor{red}{(+0.0251)}
&\cellcolor{\third}28.35\tiny\textcolor{red}{(+2.11)}
&\cellcolor{\second}0.9016\tiny\textcolor{red}{(+0.0607)}
&\cellcolor{\third}26.04\tiny\textcolor{red}{(+1.51)}
&\cellcolor{\second}0.7395\tiny\textcolor{red}{(+0.0500)}\cr

\bottomrule
\end{tabular}
}
%\vspace{-1mm}
\label{tab_hypernerf_comparison}
\end{table*}

\subsection{Comparisons and Results} 

\noindent \textbf{Quantitative Comparisons.} As observed in Tab~\ref{tab_neu3d_comparison}, Tab.~\ref{tab_nerf-ds}, and Tab.~\ref{tab_hypernerf_comparison}, our EvoGS consistently surpasses the baselines across all Neu3D, NeRF-DS, and HyperNeRF scenes, yielding markedly higher PSNR/SSIM and lower LPIPS. These gains substantiate the effectiveness of our temporal deformation evolution modeling and deformation-aware densification, and further attest to both the method’s real-world generalization and its plug-and-play applicability. Notably, these improvements are attained without increasing computational cost: all networks ($\bm{\Phi}$ and $\bm{\Psi}$) remain lightweight and our deformation-aware triggering decreases the number of Gaussians, delivering shorter training time and higher FPS. We also measure the GPU memory used by the deformation-state buffer and the peak GPU memory during training and evaluation. For EvoGS(Grid4D) on the $coffee\_martini$ scene of Neu3D, the deformation-state buffer only occupies 3.6 GB of GPU memory, while the peak GPU memory during training/evaluation is 16/8 GB. Furthermore, compared to FreeTimeGS \cite{wang2025freetimegs} which achieves 32.25 dB PSNR and 0.946 SSIM over Neu3D, our EvoGS variants based on Grid4D and 4DGS achieve 32.44/32.19 dB PSNR and 0.944/0.940 SSIM, respectively. These results show that EvoGS remains competitive with recent fully explicit approaches, while our method targets a complementary setting by improving temporal deformation modeling in deformation-based dynamic Gaussian frameworks and can be integrated into multiple such backbones.

\noindent \textbf{Visual Comparisons.} Fig.~\ref{fig_neu3d}, Fig.~\ref{fig_visual_nerf-ds}, Fig.~\ref{fig_visual_hypernerf}, Fig.~\ref{fig_neu2} highlight the perceptual quality of our method under diverse real-world motions. Across different real-world scenes, EvoGS exhibits fewer ghosting artifacts and sharper structural recovery. These visual improvements are attributed to our temporal deformation evolution modeling and gated densification, which together help stabilize deformation estimates and avoid amplifying noise. Furthermore, we fix the viewpoint and visualize novel views with varying time steps in Fig.~\ref{fig_consistency} and \ref{fig_consistency_broom}, which clearly presents the temporal consistency cross different time steps. The baseline (4DGS) suffers from poor temporal consistency, as evidenced by serious structure distortions at multiple time steps. This suggests that, without explicit cross-temporal constraints, directly employing an MLP to predict deformation tends to degenerate into time- and view-specific mappings rather than a temporally coherent field. In comparison, our developed temporal deformation evolution modeling helps achieve a more consistent geometry reconstruction across various time steps. Fig. \ref{fig_trajectory} also indicates that EvoGS learns a more effective dynamic Gaussian representation, where more Gaussians remain trackable in motion-critical areas. Although some residual floating artifacts can still be observed around the plate in Fig.~\ref{fig_visual_nerf-ds} (EvoGS(4DGS), EvoGS(MAGS)), this is mainly due to the highly degraded geometry produced by the underlying baselines (4DGS, MAGS), where the object structure is not reliably reconstructed. Our temporal deformation evolution modeling substantially improves the reconstruction by recovering more coherent object geometry and more stable appearance over time, rather than explicitly performing outlier removal or floater suppression. Therefore, a small number of residual floaters may remain in challenging regions, but the overall geometry, boundary consistency, and temporal coherence are clearly improved compared with the original baselines.

\subsection{Ablation Study}
To evaluate the effectiveness of our proposed EvoGS framework, we conduct a series of ablation experiments on the NeRF-DS dataset, considering several configurations: 1) replacing the temporal deformation evolution process with simple observation MLP (Baseline); 2) only utilizing the observation network with tri-plane spatial and B-spline temporal encoders (discussed in Sec.~\ref{sec:deform_correction}); 3) replacing the persistent deformation state buffer by querying the observation MLP with different timesteps; 4) utilizing the temporal deformation evolution process (w/o deformation-aware densification); 5) adding the uncertainty-aware densification triggering mechanism in Eq.~\ref{eq:densi_trigger} to 4); 6) adding our designed deformation-aware clone and split strategy in Eq.~\ref{eq:deformation-clone} to 5). The average quantitative results are reported in Tab.~\ref{tab_abla}. 

\begin{table}[htbp]
\centering
\setlength{\abovecaptionskip}{2mm}
\scriptsize
\setlength{\tabcolsep}{1pt}
\caption{Ablation results on NeRF-DS dataset (Baseline: D-3DGS). Removing any key components from our full EvoGS leads to reduced performance, highlighting the positive contribution of each module in our design.
}
\resizebox{0.45\textwidth}{!}{%
\centering
\begin{tabular}{l|ccc}
    \toprule
    Configurations & PSNR $\uparrow$ & SSIM $\uparrow$ & LPIPS $\downarrow$ \\
    \midrule
    1) Baseline (D-3DGS) & 23.68 & 0.8422  & 0.1983 \\
    2) Tri-plane and B-spline & 23.99  & 0.8520  & 0.1747 \\
    3) temporal evolution without buffer & 24.14 & 0.8535  & 0.1728 \\
    3) temporal evolution & 24.55 & 0.8632  & 0.1697 \\
    4) temporal evolution+Eq.~\ref{eq:densi_trigger} & 24.76 & 0.8659  & 0.1662 \\
    5) temporal evolution+Eq.~\ref{eq:densi_trigger}, Eq.~\ref{eq:deformation-clone}  & 25.04 & 0.8678 & 0.1632  \\ 
    \bottomrule
\end{tabular}
}
\label{tab_abla} 
\vspace{-3mm}
\end{table}

When temporal deformation evolution modeling is fully removed, the deformations at different time steps are determined by the observation MLP. Due to the lack of motion modeling across time, compared to the average performance of our method presented in Tab.~\ref{tab_nerf-ds}, this setting leads to a marked decrease in PSNR and LPIPS. When the tri-plane spatial and B-spline temporal encoders are equipped with the observation MLP, we observe enhanced performance (+0.3 PNSR) over the baseline. The comparison between 3) and 4) demonstrates the positive contributions of the deformation state buffer. Furthermore, compared with the original densification triggering or clone $\&$ split strategy, the improved performances in configurations 5) and 6) indicate that our proposed deformation-aware densification not only stabilizes the optimization but also enhances the representational fidelity of dynamic scenes. We additionally compare prediction-only (K=0), fixed blending (K=0.5), observation-only (K=1), and the learned adaptive correction, with all other settings kept unchanged. The learned adaptive correction achieves 25.04 dB PSNR / 0.868 SSIM, followed by fixed blending (24.64 dB / 0.863), while prediction-only performs worst (23.52 dB / 0.842). These results show that neither relying solely on historical prediction nor using a fixed fusion rule is optimal, and that dynamically balancing the historical prior and current observation provides the best performance.

\begin{figure}[!h]
    \centering
    \setlength{\abovecaptionskip}{2mm}
    \includegraphics[width=0.44\textwidth]{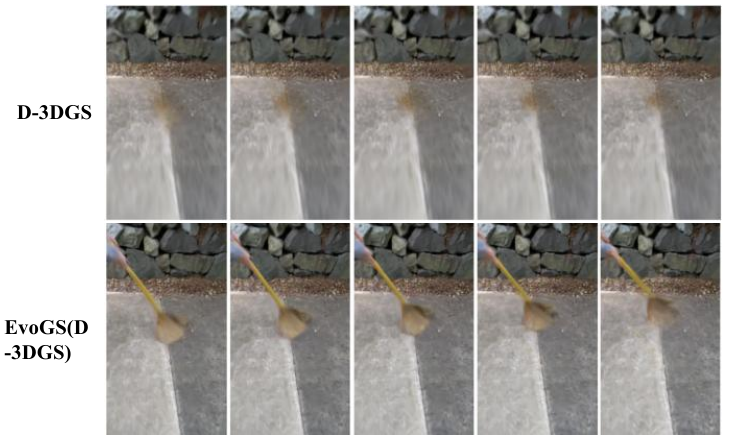}
    \caption{Visualizations of novel views rendered with the same camera pose across consecutive timesteps on the HyperNeRF (vrig) dataset. Specifically, the baseline D-3DGS fails to represent the moving broom. In contrast, EvoGS achieves good reconstructions of the moving broom and background structures across frames.}
    \label{fig_consistency_broom}
    \vspace{-3mm}
\end{figure}

\section{Conclusion and Limitation}
\label{sec_conclusion}
\noindent \textbf{Conclusion.} 
In this paper, we present EvoGS, a 3DGS-based dynamic reconstruction framework that models Gaussian deformation as a temporal evolution process. Instead of independently estimating deformations at each timestamp, EvoGS maintains persistent deformation states for each Gaussian and extrapolates the current deformation from historical corrected states. The extrapolated prior is then adaptively refined with an MLP-derived observation using temporal residual memory and motion evolution statistics. Furthermore, we introduce a deformation-aware densification mechanism that performs clone and split operations along corrected deformation directions while suppressing unstable Gaussians with uncertain motion histories. Extensive experiments demonstrate that EvoGS improves dynamic novel view synthesis quality and achieves competitive performance across multiple benchmarks.

\noindent \textbf{Limitation.} Our design targets short-horizon updates under approximately uniform frame spacing. Our formulation focuses on short-horizon deformation evolution and may become less reliable under very long occlusions, where errors in the historical states can accumulate before reliable observations become available again. Severe topology changes also remain constrained by the representational capability of the underlying deformation-based backbone.

%-------------------------------------------------------------------------

%-------------------------------------------------------------------------
% bibtex
\bibliographystyle{eg-alpha-doi} 
\bibliography{egbibsample}       

% biblatex with biber
% \printbibliography                

\newpage

\end{document}